\documentclass{article}

\PassOptionsToPackage{numbers,compress}{natbib}

\usepackage[preprint]{neurips_2025}

\usepackage[utf8]{inputenc} 
\usepackage[T1]{fontenc}    
\usepackage{url}            
\usepackage{xcolor}         

\definecolor{citegray}{RGB}{100,100,100}

\usepackage{hyperref}       
\usepackage{hypernat}       

\hypersetup{
    colorlinks=true,
    linkcolor=blue,
    citecolor=citegray,  
    urlcolor=blue,
    pagebackref=false,
    hypertexnames=false
}
\usepackage{booktabs}       
\usepackage{amsfonts}       
\usepackage{nicefrac}       
\usepackage{microtype}      

\usepackage{enumitem} 
\usepackage{float}
\usepackage{wrapfig}
\usepackage{algorithm}
\usepackage{algorithmic}
\usepackage{multirow}
\usepackage{colortbl}
\usepackage{tikz}
\usetikzlibrary{tikzmark}
\usepackage[most]{tcolorbox}

\colorlet{Red}{red!10!white}

\newtcolorbox{mybox}[1][]{
    colback=gray!5!white,
    colframe=gray!75!black,
    fonttitle=\bfseries,
    title=#1,
    enhanced,
    boxrule=0.5pt,
    left=6pt,
    right=6pt,
    top=4pt,
    bottom=4pt,
    before skip=0.2\baselineskip,
    after skip=0.2\baselineskip
}

\definecolor{myred}{rgb}{0.7, 0.3, 0.0}
\definecolor{myblue}{HTML}{054488}
\definecolor{mygreen}{HTML}{056b34}
\definecolor{mypurple}{HTML}{7B1FA2}
\newcommand{\mytextbox}[2]{\tikzmarknode[draw=#1,thick,inner sep=2pt,anchor=base]{test}{\small #2}}

\newcommand{\red}[1]{\mytextbox{myred}{\textbf{\textcolor{myred}{#1}}}}
\newcommand{\blue}[1]{\mytextbox{myblue}{\textbf{\textcolor{myblue}{#1}}}}
\newcommand{\green}[1]{\mytextbox{mygreen}{\textbf{\textcolor{mygreen}{#1}}}}
\newcommand{\purple}[1]{\mytextbox{mypurple}{\textbf{\textcolor{mypurple}{#1}}}}

\begin{document}

\title{Dep-Search: Learning Dependency-Aware Reasoning Traces with Persistent Memory}

\author{%
Yanming Liu$^{1}$\thanks{Equal contribution.} , Xinyue Peng$^{2*}$, Zixuan Yan$^{1}$, Yanxin Shen$^{1}$, Wenjie Xu$^{3}$, \\ \bf
Yuefeng Huang$^{1}$, Xinyi Wang, Jiannan Cao$^{4}$, Jianwei Yin$^{1}$, Xuhong Zhang$^{1}$\thanks{Corresponding author.} \\
$^{1}$Zhejiang University, $^{2}$Intel Corporation, $^{3}$Tsinghua University, \\
$^{4}$Massachusetts Institute of Technology \\
\texttt{{\{oceann24, yanzixuan, ssyysyx, zhangxuhong, zjuyjw\}@zju.edu.cn}}\\
\texttt{{\{xuwj24\}@mails.tsinghua.edu.cn, jiannan@mit.edu, xinyue.peng@intel.com}} \\
}

\maketitle

\begin{abstract}
Large Language Models (LLMs) have demonstrated remarkable capabilities in complex reasoning tasks, particularly when augmented with search mechanisms that enable systematic exploration of external knowledge bases. The field has evolved from traditional retrieval-augmented generation (RAG) frameworks to more sophisticated search-based frameworks that orchestrate multi-step reasoning through explicit search strategies. However, existing search frameworks still rely heavily on implicit natural language reasoning to determine search strategies and how to leverage retrieved information across reasoning steps. This reliance on implicit reasoning creates fundamental challenges for managing dependencies between sub-questions, efficiently reusing previously retrieved knowledge, and learning optimal search strategies through reinforcement learning. To address these limitations, we propose Dep-Search, a dependency-aware search framework that advances beyond existing search frameworks by integrating structured reasoning, retrieval, and persistent memory through GRPO. Dep-Search introduces explicit control mechanisms that enable the model to decompose questions with dependency relationships, retrieve information when needed, access previously stored knowledge from memory, and summarize long reasoning contexts into reusable memory entries. Through extensive experiments on seven diverse question answering datasets, we demonstrate that Dep-Search significantly enhances LLMs' ability to tackle complex multi-hop reasoning tasks, achieving substantial improvements over strong baselines across different model scales.
\end{abstract}

\section{Introduction}

Large Language Models (LLMs) have demonstrated remarkable capabilities in complex reasoning tasks, particularly when augmented with search mechanisms that enable systematic exploration of external knowledge bases \cite{lewis2020retrieval, jiang2023active, asai2024selfrag}. The field has evolved from traditional retrieval-augmented generation (RAG) frameworks to more sophisticated search-based frameworks that orchestrate multi-step reasoning through explicit search strategies \cite{li-etal-2025-search}. Recent work such as Search-R1 \cite{jin2025searchr}, DeepResearcher \cite{zheng2025deepresearcher}, Chain-of-Agents \cite{li2025chain}, and Kimi-K2 \cite{team2025kimi} have shown that search frameworks can effectively decompose complex questions, retrieve relevant information from multiple sources, and synthesize answers through structured multi-step reasoning \cite{wang2025astute}. However, existing search frameworks still rely heavily on implicit natural language reasoning to determine search strategies and how to leverage retrieved information across reasoning steps \cite{wang2025astute}. This reliance on implicit reasoning creates fundamental challenges for managing dependencies between sub-questions, efficiently reusing previously retrieved knowledge, and learning optimal search strategies through reinforcement learning \cite{zhouleast}.

The fundamental problem lies in the lack of explicit dependency modeling and persistent memory management in current search frameworks. Existing approaches decompose questions into sub-questions but fail to explicitly model dependencies between these sub-questions, leading to inefficient search patterns where the same information may be retrieved multiple times or sub-questions are answered out of dependency order \cite{liu2024ra}. Moreover, existing systems treat each reasoning episode independently, discarding valuable knowledge extracted during search that could be reused across questions or even within the same multi-step reasoning process \cite{xu2025mem}. This knowledge loss is particularly problematic in complex scenarios where retrieved facts from early steps are needed in later dependent steps, forcing redundant searches and increasing computational costs \cite{lu2025arpo}. Additionally, training search-based LLMs to learn optimal search strategies remains challenging, as existing reinforcement learning approaches struggle with the sparse reward signals and the need to jointly optimize decomposition, retrieval, memory access, and reasoning behaviors \cite{dong2025agentic}.

To address these limitations, we propose \textbf{Dep-Search}, a dependency aware search framework that advances beyond existing search frameworks by integrating structured reasoning, retrieval, and persistent memory through GRPO. Dep-Search introduces explicit control mechanisms that enable the model to decompose questions with dependency relationships, retrieve information when needed, access previously stored knowledge from memory, and summarize long reasoning contexts into reusable memory entries. By combining dependency aware question decomposition with a persistent memory system and GRPO for trajectory-level learning, Dep-Search ensures that reasoning follows explicit dependency structures, retrieved knowledge is efficiently stored and reused, and the policy learns to optimize the entire search-reasoning-memory pipeline jointly. Unlike existing search frameworks that rely on heuristic search strategies, Dep-Search treats all tokens uniformly in the policy, enabling end-to-end learning of when to decompose, what to retrieve, when to access memory, and how to synthesize final answers, while the explicit memory state provides verifiable knowledge accumulation throughout the reasoning process.

\textbf{Our Contributions.} Our contributions are detailed as follows. \begin{itemize}[leftmargin=*] 
\item We present Dep-Search, a novel framework that formalizes multi-hop reasoning through dependency aware decomposition and explicit control tokens, providing structured reasoning traces and efficient knowledge reuse.
\item We introduce a persistent memory system that automatically stores summarized facts from searches and enables efficient memory access through embedding-based similarity search, addressing the knowledge loss problem in existing search frameworks.
\item We demonstrate that QDMR-based decomposition enables adaptive dependency modeling that significantly outperforms sequential decomposition approaches, allowing the model to determine both the number of reasoning steps and their dependency structure dynamically.
\end{itemize}

\section{Related Work}

\subsection{Agentic Reinforcement Learning}
Recent advances in agentic reinforcement learning (RL) have explored how LLM-based agents can interact with environments, tools, and external knowledge sources to solve complex tasks through trial-and-error learning \cite{shao2024deepseekmath, shao2025deepseekmath, jin2025search}. Early work focused on using RL to fine-tune language models on synthetic reasoning tasks or instruction-following benchmarks, typically with short-horizon rewards and limited interaction structure \cite{ouyang2022training, song-etal-2025-smart}. Recent frameworks introduce multi-step decision processes in which the agent can iteratively call tools, plan, and revise its strategy \cite{li2025chain, team2025kimi}. These systems demonstrate that explicit interaction loops and environment feedback can significantly improve the robustness and adaptability of LLMs on complex tasks such as web navigation, code generation, and multi-hop question answering \cite{zhai2024fine, guo2024transagent}. A common theme across these approaches is the use of policy optimization  balance exploration and exploitation, such as entropy-balanced objectives that encourage diverse exploration while maintaining exploitation of promising strategies \cite{dong2025agentic}, and experience replay mechanisms that enable agents to learn from past trajectories more effectively \cite{lu2025arpo}. The emphasis on trajectory-level learning, where agents learn to optimize sequences of actions rather than individual decisions, enabling better credit assignment and long-term planning in complex multi-step reasoning scenarios. 

\subsection{Agentic Memory}
A growing line of work studies how LLM agents can maintain and exploit persistent memory to improve long-term coherence, personalization, and knowledge reuse \cite{packer2023memgpt, xu2025amem, qin2024mooncake}. Early memory-augmented systems typically log past interactions or retrieved documents in a buffer and naively prepend them to the prompt, which quickly becomes inefficient and noisy as the context grows \cite{zhang2025multiagent, gao2023retrieval}. Subsequent approaches introduce memory retrieval modules based on dense embeddings, enabling agents to select relevant past experiences or facts conditioned on the current query \cite{wang2024wise, tan2025prospect}. Recent agentic memory frameworks go further by allowing agents to write structured summaries into memory, compressing long trajectories into reusable high-level knowledge that can be recalled \cite{wang2025karma, hu2025memory, wei2025evo}. A common evolution across these approaches is the shift from passive memory storage to active memory management, where agents not only retrieve but also strategically write and organize memory content to optimize knowledge reuse across different reasoning episodes. 

\section{Methodology}

\subsection{Problem Overview}

Let the problem distribution be $\mathcal{D}$, where each instance is a natural-language question $Q \sim \mathcal{D}$. Our goal is to generate an answer $A$ through a dependency aware search process by maximizing the expected trajectory return:
\begin{equation}
\max_{\theta}\ \mathbb{E}_{Q \sim \mathcal{D},\ \tau \sim \pi_\theta(\cdot \mid Q)}\big[ R(\tau) \big],
\end{equation}
where $\pi_\theta$ is the Dep-Search policy, $\tau=(a_1,\dots,a_T)$ is a complete reasoning trajectory containing intermediate actions and the final answer, and $R(\tau)$ is the trajectory-level return.

At step $t$, the search state is defined as
\begin{equation}
S_t=(\mathcal{T}_t,\ \mathcal{C}_t,\ \mathcal{M}_t),
\end{equation}
where $\mathcal{T}_t$ is the current dependency aware reasoning trace, recording decomposed sub-questions and their dependency relations; $\mathcal{C}_t$ is the current context, including the system prompt, the question $Q$, previously generated text, retrieved evidence, and the explicitly exposed memory content; and $\mathcal{M}_t$ is the memory buffer that stores fact sentences extracted during reasoning.

In implementation, the state $S_t$ is encoded in the already generated token prefix $x_{1:t-1}$. The policy defines the conditional distribution of the next token on this prefix:
\begin{equation}
p_\theta(a_t \mid S_t)\ \equiv\ p_\theta(a_t \mid x_{1:t-1}).
\end{equation}

\subsection{Data Collection and Design}

During data collection, we use the current policy $\pi_{\theta_{\text{old}}}$ to interact with the environment and sample multiple complete reasoning trajectories for each question $Q$, which are then used for GRPO optimization.

Given a question $Q$, the initial state is
\begin{equation}
S_0=(\mathcal{T}_0,\ \mathcal{C}_0,\ \mathcal{M}_0),
\end{equation}
where $\mathcal{T}_0$ is empty, $\mathcal{C}_0$ consists of the system prompt and $Q$, and $\mathcal{M}_0$ is the fixed initial memory. The conditional probability of a full trajectory $\tau=(a_1,\dots,a_T)$ is:
\begin{equation}
p_{\theta_{\text{old}}}(\tau\mid Q)=\prod_{t=1}^{T}\pi_{\theta_{\text{old}}}(a_t\mid x_{1:t-1}),
\end{equation}
where $x_{1:t-1}$ denotes the token prefix before step $t$, and the state sequence $S_t$ is induced by the environment transition operator $S_{t+1}=\mathcal{G}(S_t,a_t)$ that updates the state based on the generated token $a_t$. Specifically, when the model emits control tokens, the environment updates the state components as follows: (1) \text{\green{<Decompose>}} updates $\mathcal{T}_t$ by adding new sub-questions and their dependency edges; (2) \text{\blue{<Retrieve>}} updates $\mathcal{C}_t$ by appending retrieved documents and automatically summarizes them into memory entries; (3) \text{\red{<Memory>}} updates $\mathcal{C}_t$ by appending retrieved memory facts; (4) \text{\purple{<Conclusion>}} updates $\mathcal{M}_t$ by summarizing the current context into new memory entries. For regular reasoning tokens, $\mathcal{C}_t$ is updated by appending the generated tokens to the context. The complete rollout procedure is detailed in Algorithm~\ref{alg:dep_search}.

\begin{figure}[t]
    \centering
    \resizebox{\textwidth}{!}{\includegraphics{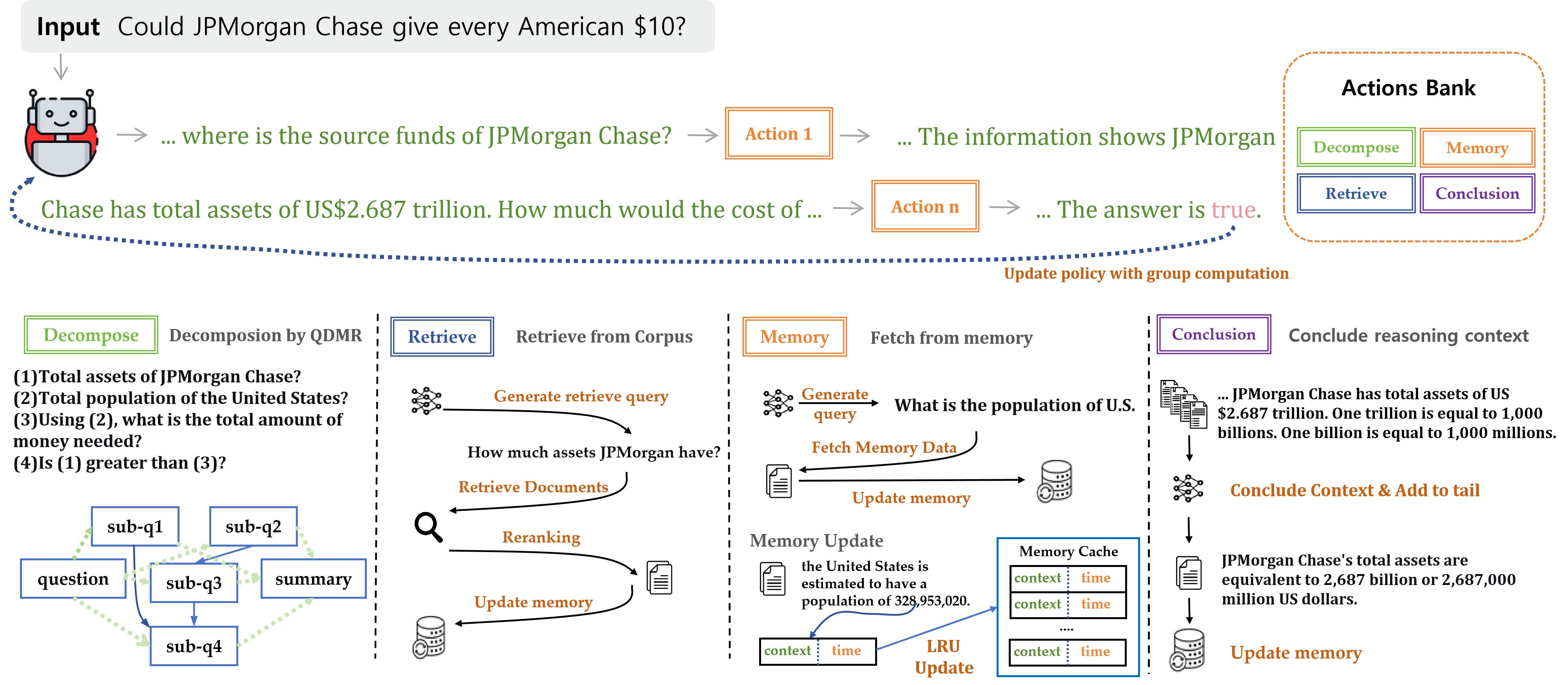}}
    \caption{Overview of the Dep-Search framework. The agent decomposes questions into dependent sub-questions, retrieves relevant information, accesses stored knowledge from memory, and synthesizes answers through trajectory-level reinforcement learning.}
    \label{fig:main}
    \end{figure}

\textbf{Decompose token.} When the model emits \text{\green{<Decompose>}}, it decomposes the question $Q$ into $K$ dependent sub-questions $\{q_1,\dots,q_K\}$, where dependencies between sub-questions form a directed acyclic graph (DAG) structure. Unlike sequential decomposition that processes sub-questions linearly, Dep-Search models explicit dependency relationships where each sub-question $q_k$ may depend on the results of one or more prerequisite sub-questions, forming a multi-level tree-like dependency structure. The model then solves these sub-questions following a topological ordering, ensuring that prerequisite sub-questions are resolved before dependent ones, similar to QDMR decomposition strategies. The decomposition updates the reasoning trace as $\mathcal{T}_{t+1} = \mathcal{T}_t \cup \{(q_k, \text{deps}(q_k))\}$, where $\text{deps}(q_k)$ denotes the set of prerequisite sub-questions for $q_k$. The detailed training prompt template is provided in Appendix~\ref{app:prompt_template}.

\textbf{Retrieve token.} When the model emits \text{\blue{<Retrieve>}} followed by a query $r_{1:L}$ and the closing tag \text{\blue{</Retrieve>}}, the environment immediately performs retrieval. The retrieval process consists of two stages: first using \textbf{qwen3-embedding} for dense retrieval to obtain a candidate set $\mathcal{D}_{\text{cand}}$ with similarity scores $s_{\text{dense}}(d_i, r) = \text{cosine}(\mathbf{E}_{\text{emb}}(d_i), \mathbf{E}_{\text{emb}}(r))$, then applying \textbf{qwen3-reranker} for re-ranking with scores $s_{\text{rerank}}(d_i, r)$ to select the top-$k$ documents:
\begin{equation}
\mathcal{D}_t = \text{Top-$k$}(\mathcal{D}_{\text{cand}}, s_{\text{rerank}}),
\end{equation}
where $\mathbf{E}_{\text{emb}}$ and $\mathbf{E}_{\text{rerank}}$ denote the embedding functions for dense retrieval and reranking, respectively. The retrieved documents are formatted and inserted as \text{\blue{<Retrieve\_result>}}$\mathcal{D}_t$\text{\blue{</Retrieve\_result>}} immediately after the closing \text{\blue{</Retrieve>}} tag, updating the context as $\mathcal{C}_{t+1} = \mathcal{C}_t \cup \mathcal{D}_t$. The model generates the query autonomously, allowing it to determine what information to retrieve based on the current reasoning context. Additionally, information from the retrieved documents is automatically summarized into fact sentences and stored in memory using the LRU rule, updating $\mathcal{M}_{t+1}$ accordingly.

\textbf{Memory token.} When the model emits \text{\red{<Memory>}} followed by a query and the closing tag \text{\red{</Memory>}}, the environment fetches relevant summarized facts from the memory buffer. Concretely, it always includes the most recently written memory items and augments them with additional entries retrieved by running qwen3-embedding over all $m\in\mathcal{M}_t$, computing cosine similarities between the query and memory embeddings, and selecting those whose similarity exceeds a threshold. The concatenated memory snippets are then inserted as \text{\red{<Memory\_result>}}$\mathcal{M}^{\mathrm{read}}_t$\text{\red{</Memory\_result>}} immediately after the closing \text{\red{</Memory>}} tag, updating the context as $\mathcal{C}_{t+1} = \mathcal{C}_t \cup \mathcal{M}^{\mathrm{read}}_t$. The model generates the query autonomously, allowing it to determine what knowledge to fetch from memory based on the current reasoning needs, enabling effective knowledge reuse during reasoning.

\textbf{Conclusion token.} When the model emits \text{\purple{<Conclusion>}}, it asks the environment to summarize the preceding reasoning and retrieved evidence that have been resolved into a compact natural-language conclusion. The environment runs the LLM in summarization mode over the current context $\mathcal{C}_t$ and writes the resulting fact sentences into the memory buffer as new entries, updating $\mathcal{M}_{t+1}$ via the LRU rule. These stored facts can later be accessed via \text{\red{<Memory>}} to avoid redundant retrieval and reasoning. Note that memory entries are not deleted but accumulated, with older entries evicted only when the memory capacity is exceeded.

\subsection{Memory Module}

We model the memory $\mathcal{M}_t$ as an LRU buffer with a fixed capacity, storing fact sentences extracted from retrieved documents and summarized reasoning during the search process. The memory and its update rule have no trainable parameters and are treated as part of the state and the environment transition.

\textbf{State representation.} At time step $t$, the memory is a finite set:
\begin{equation}
\mathcal{M}_t=\{m^{(t)}_1,\dots,m^{(t)}_{n_t}\},\quad n_t\le C_{\text{mem}}.
\end{equation}
Each entry $m$ contains a fact sentence $f(m)$ expressed in natural language and metadata such as source and write time. For example, memory entries may contain fact sentences like ``Beijing hosted the 2022 Winter Olympics'' or ``Tom Hanks starred in Forrest Gump, which grossed \$678 million worldwide.'' These fact sentences are stored as plain text, allowing the model to reuse previously extracted knowledge without re-retrieving or re-reasoning. The overall search state is
\begin{equation}
S_t=(\mathcal{T}_t,\ \mathcal{C}_t,\ \mathcal{M}_t).
\end{equation}
In practice, $\mathcal{M}_t$ is verbalized into a short ``known facts'' segment and injected into the token prefix so that the policy can explicitly read the current memory.

When the accumulated context becomes long or contains several resolved sub-questions, the model can emit \text{\purple{<Conclusion>}} to compress the preceding reasoning and evidence that have been resolved into new memory entries. We denote the set of newly written memory items at step $t$ by
\begin{equation}
\mathcal{F}_t=\text{Summarize}(\mathcal{C}_t),
\end{equation}
where $\text{Summarize}(\cdot)$ is implemented by prompting the LLM to produce a few natural-language fact sentences that capture reusable knowledge from the resolved reasoning steps. These fact sentences are then stored in $\mathcal{M}_t$ without deleting existing entries, allowing knowledge accumulation throughout the reasoning process.
We maintain a recency marker $\ell_t(m)\in\mathbb{N}$ for each memory entry $m$, and update it whenever $m$ is written/added by $\mathcal{F}_t$. Let the candidate set be:
\begin{equation}
\tilde{\mathcal{M}}_{t+1}=\mathcal{M}_t\cup \mathcal{F}_t,
\end{equation}
and update the recency markers:
\begin{equation}
\ell_{t+1}(m)=
\begin{cases}
t, & m\in \mathcal{F}_t,\\
\ell_t(m), & m\in \mathcal{M}_t\setminus \mathcal{F}_t.
\end{cases}
\end{equation}
Then we perform capacity truncation to obtain the updated memory:
\begin{equation}
\mathcal{M}_{t+1}=
\underset{\substack{\mathcal{M}\subseteq \tilde{\mathcal{M}}_{t+1}\ |\mathcal{M}|\le C_{\text{mem}}}}{\arg\max}
\ \sum_{m\in \mathcal{M}}\ell_{t+1}(m).
\end{equation}
Equivalently, we keep up to $C_{\text{mem}}$ entries with the largest $\ell_{t+1}(m)$ and evict those with the smallest $\ell_{t+1}(m)$.

During training, each episode starts from a fixed initial memory $\mathcal{M}_0$, which is typically empty, and memory is not shared across episodes. During inference, one may inject a cross-question long-term memory as $\mathcal{M}_0$ to evaluate knowledge accumulation.
\subsection{RL with Dep-Search}

In the reinforcement learning stage, we optimize the Dep-Search policy $\pi_\theta$ using GRPO. The policy generates trajectories $\tau=(a_1,\dots,a_T)$ that interleaves control tokens with reasoning tokens, where all tokens are modeled uniformly.

We use trajectory-level rewards since the final answer quality depends on the entire reasoning process. For each question $Q$, we sample $K$ trajectories:
\begin{equation}
\mathcal{G}(Q)=\{\tau_1,\dots,\tau_K\},\quad \tau_k\sim \pi_{\theta_{\text{old}}}(\cdot\mid Q),
\end{equation}
and compute trajectory-level returns $R(\tau_k)$. We define group-relative advantages:
\begin{equation}
A(\tau_k)=R(\tau_k)-\bar R(Q),\quad \bar R(Q)=\frac{1}{K}\sum_{k=1}^{K}R(\tau_k),
\end{equation}
which naturally handle varying question difficulty by comparing trajectories within the same group.

Let the token sequence of $\tau_k$ be $\{(s_{k,t},a_{k,t})\}_t$, where $s_{k,t}$ encodes the current state $S_t$. We update the policy using the clipped GRPO objective:
\begin{equation}
\begin{split}
\mathcal{L}_{\text{GRPO}}(\theta)={}&\mathbb{E}_{Q \sim \mathcal{D},\ \tau_k \sim \pi_{\theta_{\text{old}}}(\cdot\mid Q),\ t}\Big[\min\Big(\rho_{k,t}(\theta)\cdot A(\tau_k),\ \text{clip}(\rho_{k,t}(\theta),1-\epsilon,1+\epsilon)\cdot A(\tau_k)\Big)\Big]\\
&-\beta\cdot\mathbb{E}_{Q \sim \mathcal{D},\ \tau_k \sim \pi_{\theta_{\text{old}}}(\cdot\mid Q),\ t}\Big[\text{KL}\big(\pi_{\theta_{\text{old}}}(\cdot\mid s_{k,t})\ \|\ \pi_\theta(\cdot\mid s_{k,t})\big)\Big],
\end{split}
\end{equation}
where $\rho_{k,t}(\theta)=\pi_\theta(a_{k,t}\mid s_{k,t})/\pi_{\theta_{\text{old}}}(a_{k,t}\mid s_{k,t})$. Since $A(\tau_k)$ is shared across all tokens in $\tau_k$, this enables joint optimization of decomposition, retrieval, memory access, and reasoning behaviors.

The policy $\pi_\theta(a_t\mid x_{1:t-1})$ always conditions on the current state $S_t$: it decides whether to retrieve (\text{\blue{<Retrieve>}}), whether to decompose (\text{\green{<Decompose>}}), and how to generate subsequent reasoning given the available memory and evidence. Memory writing and LRU updates are purely environment rules and do not receive gradients; reinforcement learning only updates $\theta$, learning when to retrieve and how to leverage memory to achieve high-quality answers with fewer retrievals. Since memory is explicitly included in the state, Dep-Search remains a standard MDP and the dynamic memory does not violate the assumptions of GRPO.

\subsection{Reward Model}
\label{sec:reward}

The trajectory return $R(\tau)$ is primarily driven by answer quality and imposes a curved penalty on excessive retrieval and decomposition. We define $A(\tau)$ as the final answer produced by trajectory $\tau$, $A^\star$ as the gold answer, $N_{\text{ret}}(\tau)$ as the number of \text{\blue{<Retrieve>}} calls in $\tau$, and $N_{\text{dec}}(\tau)$ as the number of \text{\green{<Decompose>}} calls in $\tau$.

The total return is
\begin{equation}
R(\tau)=R_{\text{ans}}(\tau)-R_{\text{ret}}(\tau)-R_{\text{dec}}(\tau),
\end{equation}
where $R_{\text{ans}}(\tau)$ is the answer quality reward, and $R_{\text{ret}}(\tau)$ and $R_{\text{dec}}(\tau)$ are penalties for excessive retrieval and decomposition, respectively.

For the answer-quality term $R_{\text{ans}}(\tau)$, we use exact match (EM) or F1 score between the generated answer $A(\tau)$ and the gold answer $A^\star$:
\begin{equation}
R_{\text{ans}}(\tau)=\text{EM}\big(A(\tau),A^\star\big)\quad\text{or}\quad R_{\text{ans}}(\tau)=\text{F1}\big(A(\tau),A^\star\big),
\end{equation}
where both metrics are normalized to $[0,1]$.

For both retrieval and decomposition penalties, we apply a linear penalty only when the operation count exceeds a threshold. Let $k_1$ and $k_2$ be the thresholds for retrieval and decomposition, respectively. The penalty functions are:
\begin{equation}
R_{\text{ret}}(\tau)=
\begin{cases}
0, & N_{\text{ret}}(\tau)\le k_1,\\[4pt]
\lambda_{\text{ret}}\big(N_{\text{ret}}(\tau)-k_1\big), & N_{\text{ret}}(\tau)> k_1,
\end{cases}
\quad
R_{\text{dec}}(\tau)=
\begin{cases}
0, & N_{\text{dec}}(\tau)\le k_2,\\[4pt]
\lambda_{\text{dec}}\big(N_{\text{dec}}(\tau)-k_2\big), & N_{\text{dec}}(\tau)> k_2,
\end{cases}
\end{equation}
where $\lambda_{\text{ret}}>0$ and $\lambda_{\text{dec}}>0$ control the penalty slopes for retrieval and decomposition, respectively.

This design makes the return mainly driven by answer quality, while penalizing retrieval and decomposition only after surpassing reasonable thresholds, encouraging efficient dependency aware search. Moreover, within each GRPO group for a given question, trajectories are rolled out under the same initial memory configuration and environment rules, so the model observes consistent memory dynamics across the group. This shared memory context allows GRPO to provide stable supervision for learning when and how to emit \text{\red{<Memory>}} to effectively exploit stored facts.

\section{Experimental Setup}

\subsection{Datasets}

We evaluate Dep-Search on six multi-hop question answering datasets that require complex reasoning over multiple documents. \textbf{HotpotQA} \cite{yang2018hotpotqa} is a widely-used benchmark featuring questions that require reasoning over multiple Wikipedia paragraphs, with both distractor and full-wiki settings. \textbf{2WikiMultihopQA} \cite{ho2020constructing} focuses on multi-hop questions that require comparing and contrasting information from different Wikipedia articles. \textbf{Musique} \cite{trivedi2022musique} presents questions that need to aggregate information across multiple paragraphs, with explicit reasoning chains. \textbf{Bamboogle} \cite{press2023measuring} is a challenging dataset that requires searching through multiple web pages to answer questions. \textbf{TriviaQA} \cite{joshi-etal-2017-triviaqa} contains question-answer pairs with evidence from Wikipedia and web sources, testing the model's ability to retrieve and synthesize information. \textbf{PopQA} \cite{mallen2023not} focuses on popular entity questions that require up-to-date knowledge retrieval. These datasets cover diverse question types, from factoid queries to complex multi-step reasoning, providing comprehensive evaluation of Dep-Search's dependency aware search capabilities.

\subsection{Baselines}

We compare Dep-Search against ten baseline methods: Directly Inference, Vanilla RAG, IRCoT \cite{trivedi2023interleaving}, RA-ISF \cite{liu2024ra}, Search-O1 \cite{li-etal-2025-search}, Search-R1 \cite{jin2025searchr}, R1-Searcher \cite{song2025r1}, HierSearch \cite{tan2025hiersearch}, O$^2$-Searcher \cite{sun2025zerosearch}, and ZeroSearch \cite{sun2025zerosearch}. These baselines cover the spectrum from simple retrieval-augmented generation to sophisticated search-based reasoning. For more details, please refer to Appendix~\ref{app:experimental_details}.

\subsection{Models and Metrics}

We conduct experiments using two model sizes: \textbf{Qwen2.5-3B-Instruct} and \textbf{Qwen2.5-7B-Instruct} \cite{qwen2025qwen25technicalreport}. These models provide a good balance between performance and computational efficiency, allowing us to evaluate Dep-Search's effectiveness across different model scales. For evaluation metrics, we use \textbf{Exact Match (EM)} for multiple-choice questions where the answer format is constrained, and \textbf{F1 score} for open-ended questions where partial credit is appropriate. Both metrics are normalized to $[0,1]$ and align with the reward function used during training, as described in Section~\ref{sec:reward}. Detailed information about retrieval corpus and implementation details are provided in Appendix~\ref{app:experimental_details}.

\section{Experiments}

\begin{table}[t]
    \caption{Main experimental results on single-hop and multi-hop question answering datasets. \textbf{Bold} numbers indicate the best performance among all methods for each model.}
\label{tab:main_result}
\tiny
    \renewcommand{\arraystretch}{1}
	 \resizebox{\linewidth}{!}{
\begin{tabular}{lcccccccc}
\toprule
\multicolumn{1}{c}{\multirow{2.5}{*}{\textbf{Method}}} & \multicolumn{3}{c}{\textbf{Single-Hop QA}} & \multicolumn{4}{c}{\textbf{Multi-Hop QA}} & \multicolumn{1}{c}{\multirow{2.5}{*}{\textbf{Avg.}}} \\
\cmidrule(r){2-4} \cmidrule(r){5-8}
\multicolumn{1}{c}{} & \textbf{NQ} & \textbf{TriviaQA} & \textbf{PopQA} & \textbf{HotpotQA} & \textbf{2WikiMHQA} & \textbf{Musique} & \textbf{Bamboogle} & \multicolumn{1}{c}{} \\
\midrule

\multicolumn{9}{l}{\textbf{\textit{Qwen2.5-3B-Instruct}}} \\ \midrule
Directly Inference & 12.40 & 30.60 & 12.40 & 16.00 & 19.20 & 4.40 & 16.80 & 16.00 \\
Vanilla RAG & 13.80 & 29.20 & 14.60 & 13.40 & 17.20 & 3.20 & 14.40 & 15.11 \\
IRCoT & 14.20 & 34.80 & 20.80 & 19.60 & 28.40 & 6.40 & 5.56 & 18.54 \\
RA-ISF & 15.60 & 36.20 & 22.40 & 20.80 & 29.60 & 7.20 & 6.20 & 19.71 \\
Search-O1 & 16.60 & 31.00 & 24.80 & 14.80 & 22.40 & 5.20 & 22.40 & 19.77 \\
Search-R1 & 35.80 & 55.80 & 26.00 & 33.20 & 26.00 & 7.60 & 12.50 & 28.13 \\
R1-Searcher & 37.60 & 56.20 & 32.20 & 31.20 & 29.80 & 9.40 & 18.50 & 29.85 \\
HierSearch & 44.80 & 61.00 & \textbf{48.80} & 35.00 & 34.80 & 12.40 & 22.56 & 36.31 \\
O2-Searcher & 44.20 & 60.40 & 40.40 & 34.60 & 34.40 & 12.00 & 21.44 & 35.21 \\
ZeroSearch & 36.20 & 54.40 & 25.10 & 29.00 & 28.20 & 8.80 & 16.67 & 27.54 \\
\cellcolor{Red}Dep-Search & \cellcolor{Red}\textbf{47.20} & \cellcolor{Red}\textbf{65.00} & \cellcolor{Red}47.40 & \cellcolor{Red}\textbf{38.00} & \cellcolor{Red}\textbf{38.80} & \cellcolor{Red}\textbf{14.60} & \cellcolor{Red}24.00 & \cellcolor{Red}\textbf{39.29} \\ \midrule

\multicolumn{9}{l}{\textbf{\textit{Qwen2.5-7B-Instruct}}} \\ \midrule
Directly Inference & 11.60 & 35.60 & 13.20 & 16.40 & 22.20 & 4.80 & 14.40 & 16.89 \\
Vanilla RAG & 13.20 & 36.80 & 15.40 & 17.60 & 23.40 & 5.60 & 15.20 & 17.60 \\
IRCoT & 27.60 & 47.40 & 27.40 & 21.00 & 29.20 & 9.80 & 27.78 & 27.17 \\
RA-ISF & 28.80 & 49.20 & 29.20 & 22.40 & 30.60 & 10.60 & 28.90 & 28.67 \\
Search-O1 & 19.40 & 40.60 & 25.60 & 17.00 & 27.00 & 8.60 & 30.40 & 24.06 \\
Search-R1 & 42.40 & 63.40 & 51.60 & 32.80 & 33.20 & 17.40 & 26.39 & 38.17 \\
R1-Searcher & 44.20 & 64.80 & 53.20 & 34.20 & 35.80 & 18.60 & 28.89 & 39.85 \\
HierSearch & 48.20 & 67.00 & \textbf{61.60} & 38.80 & 39.60 & 20.40 & 32.00 & 46.66 \\
O2-Searcher & 47.40 & 66.20 & 58.20 & 38.20 & 39.00 & 20.00 & 30.89 & 45.70 \\
ZeroSearch & 41.60 & 62.60 & 50.40 & 32.20 & 32.80 & 16.80 & 25.56 & 37.54 \\
\cellcolor{Red}Dep-Search & \cellcolor{Red}\textbf{53.80} & \cellcolor{Red}\textbf{72.00} & \cellcolor{Red}60.20 & \cellcolor{Red}\textbf{44.40} & \cellcolor{Red}\textbf{45.20} & \cellcolor{Red}\textbf{22.20} & \cellcolor{Red}30.56 & \cellcolor{Red}\textbf{49.77} \\ \bottomrule
\end{tabular} }
\end{table}

\subsection{Main Results}

Table~\ref{tab:main_result} presents the comprehensive evaluation results across six question answering datasets. Our Dep-Search method achieves the best overall performance on both model scales, demonstrating consistent improvements over existing baseline methods.

\textbf{Overall Performance.} Dep-Search achieves average scores of 39.29 and 49.77 on Qwen2.5-3B-Instruct and Qwen2.5-7B-Instruct, respectively, outperforming all baseline methods. On the 3B model, Dep-Search scores 39.29, about 3 points higher than HierSearch, which reaches 36.31, and about 4 points higher than O2-Searcher at 35.21. On the 7B model, Dep-Search reaches 49.77, improving over HierSearch at 46.66 by roughly 3 points and over O2-Searcher at 45.70 by roughly 4 points. These results demonstrate that our dependency aware decomposition, persistent memory mechanism, and GRPO-based training effectively improve multi-hop reasoning capabilities.

\textbf{Single-Hop vs. Multi-Hop QA.} Dep-Search shows strong performance across both single-hop and multi-hop question types. On single-hop datasets such as NQ, TriviaQA, and PopQA, Dep-Search achieves average scores of 53.07 and 62.00 on the 3B and 7B models, respectively. HierSearch slightly surpasses Dep-Search on PopQA, where HierSearch obtains 48.80 compared to 47.40 for Dep-Search on 3B, and 61.60 compared to 60.20 on 7B, while our method achieves the best results on NQ and TriviaQA. On multi-hop datasets including HotpotQA, 2WikiMHQA, Musique, and Bamboogle, Dep-Search achieves average scores of 28.85 and 35.59 on 3B and 7B models, showing larger gains over baselines. This suggests that explicit dependency modeling and memory reuse are particularly valuable for complex reasoning chains that require information from multiple sources.

\textbf{Model Scale Analysis.} The performance gap between 3B and 7B models highlights the importance of model capacity for complex reasoning tasks. Dep-Search improves from 39.29 on the 3B model to 49.77 on the 7B model, an increase of about 10.5 points that is larger than for most baselines, suggesting that larger models better leverage the structured reasoning and memory mechanisms. On the 7B model, Dep-Search achieves particularly strong performance on multi-hop datasets: on HotpotQA, Dep-Search scores 44.40, about 12 points higher than Search-R1 at 32.80; on 2WikiMHQA, Dep-Search reaches 45.20, about 12 points higher than Search-R1 at 33.20. Dep-Search's dependency aware decomposition allows the model to answer sub-questions in the correct order, while the persistent memory mechanism reduces redundant retrievals by storing and reusing previously extracted facts. 

\subsection{Ablation Study}

\begin{table}[ht!]
    \centering
    \caption{Ablation study on Qwen2.5-3B-Instruct across all datasets. We report scores when removing QDMR-style decomposition, the memory module, and the conclusion-based summarization mechanism. $\Delta$ denotes the average performance drop compared to the full model.}
    \label{tab:ablation_3b}
    \small
    \renewcommand{\arraystretch}{1.1}
    \resizebox{\linewidth}{!}{
    \begin{tabular}{lccccccccc}
    \toprule
    \multicolumn{1}{c}{\multirow{2.5}{*}{\textbf{Variant}}} & \multicolumn{3}{c}{\textbf{Single-Hop QA}} & \multicolumn{4}{c}{\textbf{Multi-Hop QA}} & \multicolumn{1}{c}{\multirow{2.5}{*}{\textbf{Avg.}}} & \multicolumn{1}{c}{\multirow{2.5}{*}{\textbf{$\Delta$}}} \\
    \cmidrule(r){2-4} \cmidrule(r){5-8}
    \multicolumn{1}{c}{} & \textbf{NQ} & \textbf{TriviaQA} & \textbf{PopQA} & \textbf{HotpotQA} & \textbf{2WikiMHQA} & \textbf{Musique} & \textbf{Bamboogle} & \multicolumn{1}{c}{} & \multicolumn{1}{c}{} \\
    \midrule
    \textbf{Full Dep-Search} & \textbf{47.20} & \textbf{65.00} & \textbf{47.40} & \textbf{38.00} & \textbf{38.80} & \textbf{14.60} & \textbf{24.00} & \textbf{39.29} & -- \\
    \midrule
    w/o QDMR Decompose & 43.80 & 61.20 & 44.00 & 34.00 & 35.20 & 12.40 & 21.20 & 35.97 & -3.32 \\
    w/o Memory Module & 41.60 & 58.40 & 42.20 & 32.50 & 33.00 & 11.00 & 19.60 & 34.04 & -5.25 \\
    w/o Conclusion & 45.00 & 62.80 & 45.60 & 35.50 & 36.60 & 13.20 & 22.40 & 37.30 & -1.99 \\
    \bottomrule
    \end{tabular}}
    \vspace{-0.15cm}
    \end{table}

To better understand the contribution of different components in Dep-Search, we conduct a comprehensive ablation study on the Qwen2.5-3B-Instruct model across all seven datasets. We systematically remove the QDMR-style decomposition, the memory module, and the explicit conclusion-based summarization to evaluate their individual contributions.

The results in Table~\ref{tab:ablation_3b} show that all three components contribute consistently across both single-hop and multi-hop datasets. Removing the memory module causes the largest performance drop, with an average decrease of 5.25 points. The degradation is particularly severe on multi-hop datasets such as Musique, where performance drops from 14.60 to 11.00, and Bamboogle, where it decreases from 24.00 to 19.60, demonstrating that reusing summarized facts across reasoning steps is essential for complex multi-hop reasoning. Removing QDMR-style decomposition leads to the second largest drop, with an average decrease of 3.32 points. The effects are more pronounced on multi-hop datasets such as HotpotQA and Musique, where performance decreases substantially, compared to single-hop datasets like TriviaQA, where the impact is more moderate, confirming that explicit dependency aware decomposition is crucial for structuring reasoning across sub-questions. The conclusion-based summarization mechanism contributes a smaller but consistent improvement, with an average gain of 1.99 points, suggesting that explicitly distilling long reasoning traces into compact, reusable summaries further stabilizes the search process. 

\subsection{Reward Function Threshold Analysis}

\begin{wrapfigure}{r}{0.45\textwidth}
    \centering
    \resizebox{\linewidth}{!}{\includegraphics{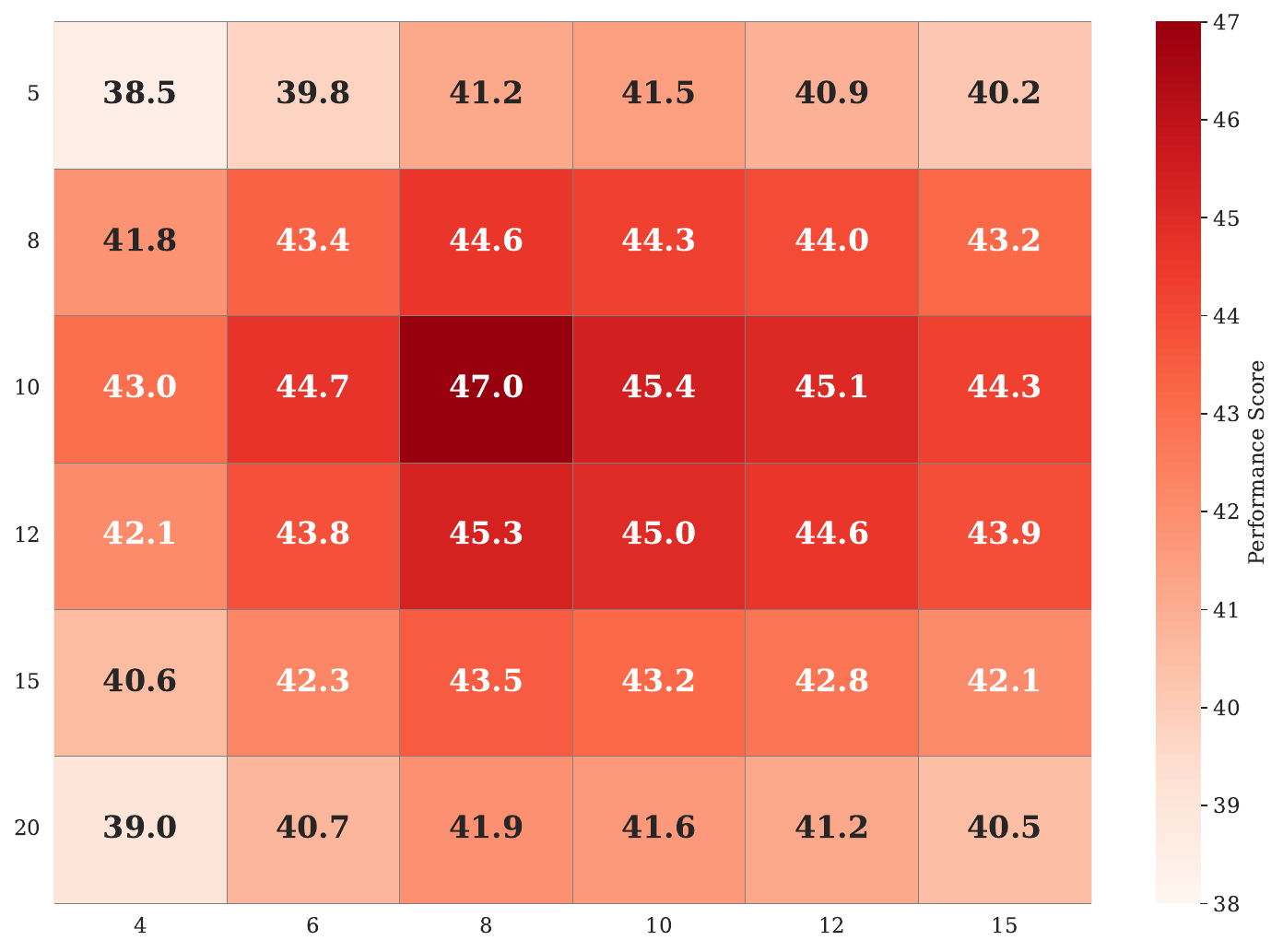}}
    \caption{Reward function threshold sensitivity analysis on 2WikiMHQA.}
    \label{fig:threshold_sensitivity}
    \end{wrapfigure}

We investigate the sensitivity of Dep-Search to the reward function thresholds $k_1$ and $k_2$  on the 2WikiMHQA dataset using Qwen2.5-7B-Instruct. These thresholds control when penalties are applied for excessive retrieval and decomposition operations, balancing between allowing necessary operations and discouraging wasteful ones.

Figure~\ref{fig:threshold_sensitivity} presents the performance across different combinations of $k_1$ and $k_2$. The optimal configuration is $k_1=10$ and $k_2=8$, achieving a score of 47.0 on 2WikiMHQA. As the retrieval threshold $k_1$ decreases, the penalty is applied earlier, discouraging necessary retrieval operations and limiting the model's ability to gather sufficient information. Conversely, as $k_1$ increases, excessive retrieval operations waste computational resources without improving answer quality. Similarly, when the decomposition threshold $k_2$ decreases, the model is penalized for necessary decomposition steps, preventing proper question breakdown. When $k_2$ increases, the model over-decomposes questions into unnecessary fine-grained steps. The optimal thresholds strike a balance that allows sufficient operations for complex multi-hop reasoning while preventing wasteful ones, demonstrating the importance of careful hyperparameter tuning for reward function design.

\subsection{Action Usage Analysis}

\begin{wrapfigure}{r}{0.5\textwidth}
    \centering
    \resizebox{\linewidth}{!}{\includegraphics{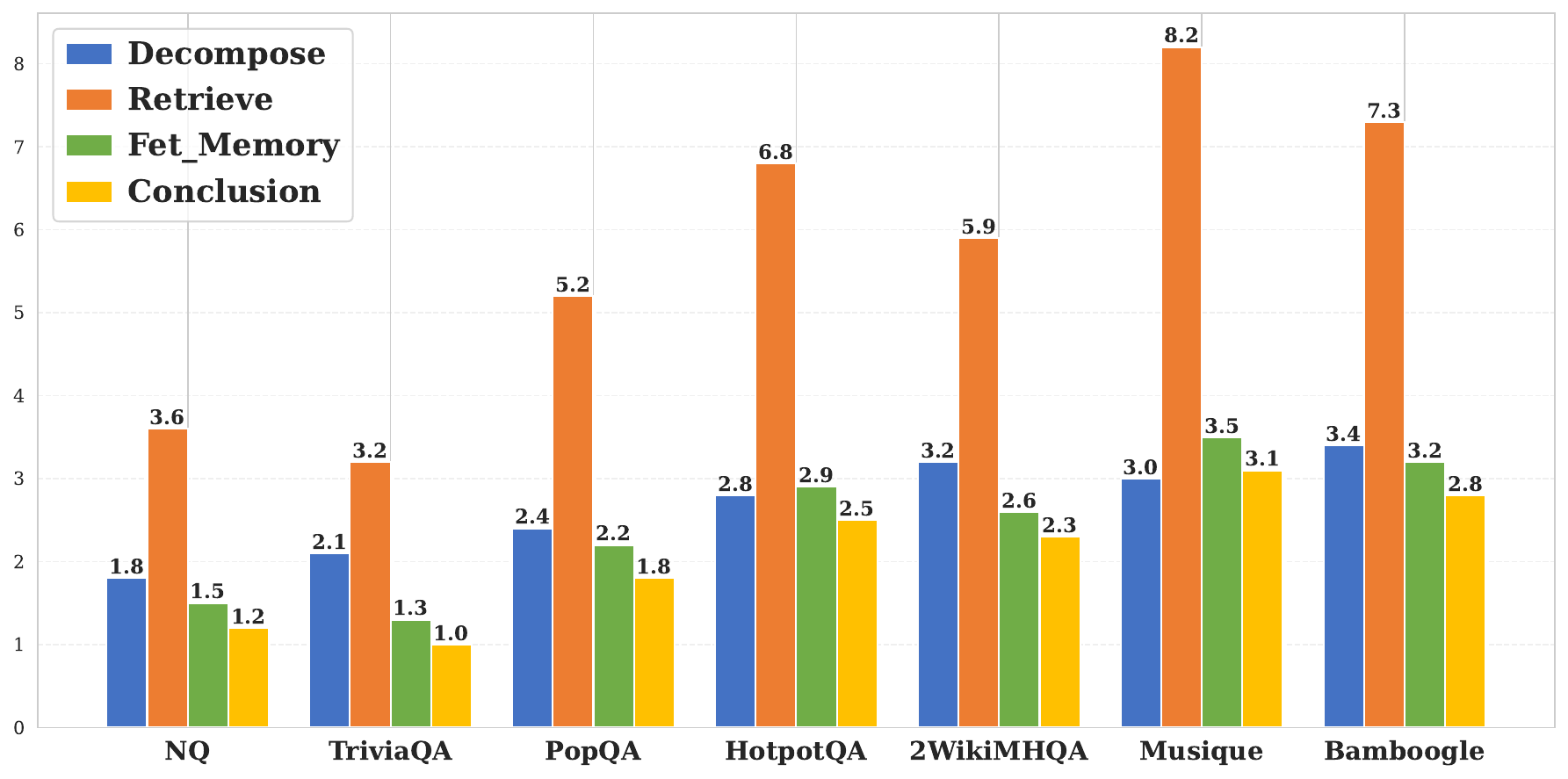}}
    \caption{Action call frequency per question across different datasets on Qwen2.5-7B-Instruct.}
    \label{fig:action_frequency}
    \end{wrapfigure}

To understand how Dep-Search adapts its search strategy to different question types, we analyze the frequency of different action calls across various datasets. This analysis reveals how the framework adjusts its decomposition, retrieval, memory access, and summarization behaviors based on dataset characteristics. Figure~\ref{fig:action_frequency} presents the average frequency of each action type across different datasets.

\textbf{Decomposition.} Multi-hop datasets trigger more frequent decomposition operations, with frequencies ranging from 1.8 to 3.4 calls per question, as the model needs to explicitly break down complex questions into dependent sub-problems. This enables the framework to structure reasoning chains with clear dependencies, allowing each sub-question to leverage results from previous steps.

\textbf{Retrieval.} Multi-hop datasets trigger extensive retrieval operations, with frequencies ranging from 3.2 to 8.2 calls per question, as the model needs to gather evidence from different documents or paragraphs to answer dependent sub-questions. The framework strategically performs retrievals at different stages of reasoning, targeting specific information needed for each step.

\textbf{Memory Access.} Memory access frequencies range from 1.3 to 3.5 calls per question, typically 40\% to 50\% of the retrieval frequency, indicating selective utilization of stored knowledge. This enables efficient knowledge reuse across reasoning chains, particularly in multi-hop scenarios where early retrieved facts are needed in later dependent steps.

\textbf{Conclusion.} Conclusion frequencies range from 1.0 to 3.1 calls per question, with multi-hop datasets showing higher frequencies as they generate longer reasoning chains that require compression. The model summarizes intermediate results into memory entries, helping manage context length and enabling knowledge reuse in subsequent reasoning steps.

\subsection{Memory Capacity Sensitivity Analysis}

\begin{wrapfigure}{r}{0.65\textwidth}
    \centering
    \resizebox{\linewidth}{!}{\includegraphics{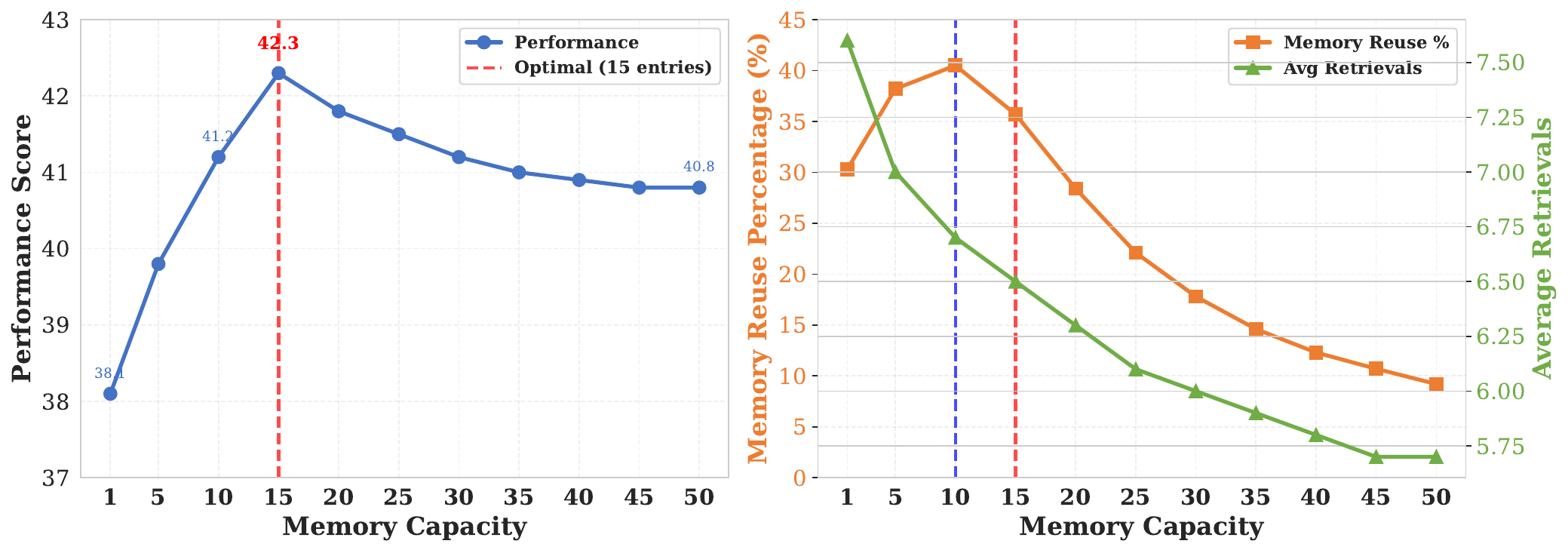}}
    \caption{Memory capacity sensitivity analysis on 2WikiMHQA.}
    \label{fig:memory_capacity}
    \end{wrapfigure}

We investigate how memory capacity affects Dep-Search performance by varying the memory buffer size from 1 to 50 entries on 2WikiMHQA using Qwen2.5-7B-Instruct. This analysis helps understand the trade-off between memory capacity and performance, identifying the optimal capacity for efficient knowledge reuse.

Figure~\ref{fig:memory_capacity} presents the performance across different memory capacities, tested at intervals of 5 entries from 1 to 50. \textbf{Performance peaks at 15 entries with a score of 42.3}, demonstrating that moderate memory capacity provides optimal knowledge reuse. Performance increases steadily from 1 to 15 entries, with scores improving from 38.1 to 42.3, as the memory buffer becomes large enough to store relevant facts without excessive overhead. However, beyond 15 entries, performance gradually decreases, dropping to 40.8 at 50 entries. This pattern suggests that while larger memory buffers can store more information, they may introduce noise or make it harder for the model to identify the most relevant entries, leading to suboptimal memory access decisions. \textbf{Memory reuse percentage peaks at 10 entries with 40.5\% of entries being reused}, then decreases rapidly as capacity increases, dropping to 9.2\% at 50 entries. This indicates that smaller capacities enable more frequent reuse of stored knowledge, while larger buffers store more one-time-use information. The optimal performance at 15 entries occurs despite lower reuse percentage compared to 10 entries, suggesting that a balance between reuse frequency and memory capacity is crucial for overall reasoning quality. The average number of retrievals decreases with larger memory capacity, suggesting that the optimal capacity balances between storing sufficient knowledge and maintaining efficient memory access. These results demonstrate that a capacity of 15 entries provides the optimal balance between performance and efficiency for 2WikiMHQA.

\section{Conclusions}

In this work, we introduced Dep-Search, a dependency aware search framework that enables LLMs to perform structured multi-hop reasoning through explicit dependency modeling and persistent memory management. Unlike existing search frameworks that rely on implicit natural language reasoning to determine search strategies, Dep-Search integrates structured reasoning, retrieval, and persistent memory through GRPO, allowing autonomous decomposition, strategic retrieval, and efficient knowledge reuse across reasoning steps. Through extensive experiments on seven diverse question answering datasets, we demonstrated that Dep-Search significantly enhances LLMs' ability to tackle complex multi-hop reasoning tasks, achieving substantial relative improvements over strong baselines across different model scales, with larger models showing greater absolute gains while smaller models benefit from more pronounced relative improvements. Our analysis also provides key insights into RL training strategies for dependency aware search-augmented reasoning, particularly regarding reward function design and the interplay between decomposition, retrieval, and memory access behaviors. Looking ahead, future work can explore expanding Dep-Search to support broader reasoning scenarios, including more sophisticated dependency modeling mechanisms, dynamic memory management strategies, and integration with diverse external knowledge sources beyond Wikipedia.

\bibliographystyle{plainnat}
\bibliography{custom}

\appendix

\section{Experimental Setup Details}
\label{app:experimental_details}

\subsection{Datasets}

We evaluate Dep-Search on six multi-hop question answering datasets that require complex reasoning over multiple documents:
\begin{itemize}[leftmargin=*]
\item \textbf{HotpotQA} \cite{yang2018hotpotqa}: A widely-used benchmark featuring questions that require reasoning over multiple Wikipedia paragraphs, with both distractor and full-wiki settings. The dataset contains over 113,000 question-answer pairs, where each question requires combining information from at least two paragraphs to answer correctly. Questions are designed to test various reasoning types including comparison, bridge, and intersection queries. The distractor setting includes irrelevant paragraphs to test the model's ability to filter noise, while the full-wiki setting requires searching through the entire Wikipedia corpus. The dataset's emphasis on multi-paragraph reasoning makes it an ideal testbed for dependency aware reasoning, as models must identify which paragraphs contain prerequisite information before answering dependent questions.

\item \textbf{2WikiMultihopQA} \cite{ho2020constructing}: Focuses on multi-hop questions that require comparing and contrasting information from different Wikipedia articles. The dataset contains over 190,000 question-answer pairs that explicitly require reasoning across multiple Wikipedia articles. Questions often involve identifying relationships between entities mentioned in different articles, such as comparing birth dates, locations, or achievements. This dataset emphasizes the need for explicit dependency modeling, as questions frequently require identifying prerequisite information from one article before querying related information from another. The structured nature of Wikipedia articles and the explicit multi-hop requirements make this dataset particularly suitable for evaluating dependency aware search frameworks.

\item \textbf{Musique} \cite{trivedi2022musique}: Presents questions that need to aggregate information across multiple paragraphs, with explicit reasoning chains. The dataset is constructed by composing simpler single-hop questions into complex multi-hop queries, resulting in over 25,000 questions. Each question comes with annotated reasoning paths that specify the sequence of information needed to answer correctly. The dataset provides explicit reasoning chains, enabling evaluation of whether models can correctly structure multi-step reasoning. Questions require aggregating facts from multiple paragraphs, often involving temporal reasoning, numerical comparisons, or logical deductions. The composition-based construction ensures that questions have clear dependency structures, making it valuable for testing dependency aware decomposition strategies.

\item \textbf{Bamboogle} \cite{press2023measuring}: A challenging dataset that requires searching through multiple web pages to answer questions. The dataset contains questions that simulate real-world web search scenarios, where answers are distributed across different web pages. Questions are designed to test the model's ability to navigate complex information spaces, follow links between pages, and synthesize information from multiple sources. The dataset emphasizes the importance of managing dependencies across different sources, as information from one page may be needed to understand or locate information on another page. This dataset tests the model's ability to handle noisy web content and manage search dependencies in unstructured information spaces.

\item \textbf{TriviaQA} \cite{joshi-etal-2017-triviaqa}: Contains question-answer pairs with evidence from Wikipedia and web sources, testing the model's ability to retrieve and synthesize information. The dataset includes over 650,000 question-answer-evidence triples, with questions authored by trivia enthusiasts. Each question comes with multiple evidence documents (approximately six per question on average), providing high-quality distant supervision. The dataset includes both reading comprehension and open-domain question answering formats, testing different aspects of retrieval and reasoning capabilities. Questions exhibit considerable syntactic and lexical variability between questions and corresponding answer-evidence sentences, requiring models to perform cross-sentence reasoning. The large scale and diverse question types make TriviaQA a comprehensive benchmark for evaluating retrieval-augmented reasoning systems.

\item \textbf{PopQA} \cite{mallen2023not}: Focuses on popular entity questions that require up-to-date knowledge retrieval. The dataset contains questions about popular entities that are frequently queried, testing the model's ability to retrieve and utilize factual knowledge. Questions often involve temporal reasoning, as they may ask about recent events or current information about well-known entities. This dataset emphasizes the importance of memory mechanisms for storing and reusing factual knowledge, as questions about the same entity may appear multiple times with different aspects. The focus on popular entities ensures that questions have sufficient context and evidence available, while still requiring sophisticated reasoning to combine multiple facts about the same entity.
\end{itemize}
These datasets cover diverse question types, from factoid queries to complex multi-step reasoning, providing comprehensive evaluation of Dep-Search's dependency aware search capabilities.

\subsection{Baselines}

We compare Dep-Search against ten baseline methods that represent different approaches to multi-hop question answering:
\begin{itemize}[leftmargin=*]
\item \textbf{Directly Inference}: Uses the base language model without any retrieval or search mechanisms, serving as a lower bound to demonstrate the importance of external knowledge access. This baseline directly generates answers from the model's parametric knowledge, without accessing external documents or performing any search operations. It helps quantify the performance gain achieved by incorporating retrieval and search mechanisms.

\item \textbf{Vanilla RAG}: Retrieves relevant documents using dense embeddings and directly generates answers from the retrieved context, representing the simplest form of retrieval-augmented generation. This baseline performs a single retrieval step using dense embeddings, retrieves top-$k$ documents, and generates answers directly from the concatenated retrieved context. It demonstrates the baseline performance achievable with simple retrieval-augmented generation without iterative reasoning or search strategies.

\item \textbf{IRCoT} \cite{trivedi2023interleaving}: Integrates iterative retrieval with chain-of-thought reasoning, retrieving documents at each reasoning step. The method alternates between generating reasoning steps and retrieving relevant documents based on the current reasoning context. This approach demonstrates the benefits of interleaving retrieval and reasoning, allowing the model to refine its search queries based on intermediate reasoning results. The iterative process enables the model to progressively gather information needed for answering complex questions.

\item \textbf{RA-ISF} \cite{liu2024ra}: Employs retrieval-augmented inference with iterative search and filtering, using feedback mechanisms to refine retrieval queries. The method performs multiple rounds of retrieval, where each round uses feedback from previous retrievals to improve query formulation. It employs iterative search and filtering mechanisms to progressively narrow down relevant information, enabling more targeted retrieval as reasoning progresses.

\item \textbf{Search-O1} \cite{li-etal-2025-search}: Integrates agentic retrieval mechanisms with large reasoning models, orchestrating multi-step reasoning through explicit search strategies. The framework combines large reasoning models with autonomous search capabilities, allowing the model to decide when and what to search based on reasoning needs. Search-O1 employs explicit search tokens and integrates retrieval results into the reasoning process, enabling coordinated search and reasoning behaviors.

\item \textbf{Search-R1} \cite{jin2025searchr}: Uses reinforcement learning to train models to reason and leverage search engines, representing a recent search-based framework for multi-step reasoning. The method trains language models to autonomously decide when to search and how to formulate search queries through reinforcement learning. Search-R1 demonstrates the effectiveness of learning search strategies through RL, enabling models to develop effective search behaviors through trial and error.

\item \textbf{R1-Searcher} \cite{song2025r1}: Implements recursive search mechanisms based on R1 architecture for complex queries, enabling deeper exploration of search spaces. The method employs recursive search strategies that allow the model to iteratively refine queries and explore search spaces more thoroughly. R1-Searcher's recursive approach enables handling of complex queries that require multiple levels of reasoning, allowing for deeper information exploration.

\item \textbf{HierSearch} \cite{tan2025hiersearch}: Employs hierarchical search strategies that decompose questions at multiple granularity levels, allowing for more structured reasoning processes. The framework decomposes questions hierarchically, creating multiple levels of abstraction that guide the search process. HierSearch's multi-granularity approach enables more structured reasoning by organizing search at different levels of detail.

\item \textbf{O$^2$-Searcher} \cite{sun2025zerosearch}: Focuses on optimizing search efficiency through advanced architectural designs. The method employs sophisticated search mechanisms designed to minimize unnecessary search operations while maintaining high answer quality. O$^2$-Searcher optimizes the trade-off between search cost and performance, enabling efficient search-augmented reasoning.

\item \textbf{ZeroSearch} \cite{sun2025zerosearch}: Aims to incentivize search capabilities without explicit search operations, representing a state-of-the-art search framework. The method trains models to internalize search behaviors without requiring explicit search API calls, encouraging the model to develop search-like reasoning patterns. ZeroSearch demonstrates an alternative approach to search-augmented reasoning by learning implicit search strategies.
\end{itemize}
These baselines cover the spectrum from simple retrieval-augmented generation to sophisticated search-based reasoning, allowing us to assess Dep-Search's improvements in dependency modeling and memory management.

\subsection{Models}

We conduct experiments using two model sizes: \textbf{Qwen2.5-3B-Instruct} and \textbf{Qwen2.5-7B-Instruct} \cite{qwen2025qwen25technicalreport}. These models provide a good balance between performance and computational efficiency, allowing us to evaluate Dep-Search's effectiveness across different model scales. Qwen2.5 is a family of large language models that demonstrate strong reasoning capabilities and instruction-following abilities. The models are trained with extensive instruction tuning and demonstrate competitive performance on various reasoning benchmarks. The 3B variant offers faster inference and lower memory requirements, making it suitable for resource-constrained environments, while the 7B variant provides stronger reasoning capabilities and better instruction understanding. Both variants support the control tokens and structured reasoning required by Dep-Search, enabling comprehensive evaluation of the framework's dependency aware search mechanisms across model scales. The choice of these model sizes allows us to demonstrate that Dep-Search's improvements are consistent across different model capacities, suggesting that the framework's benefits are not limited to larger models.

\subsection{Retrieval Corpus}

All retrieval operations are performed over the \textbf{Wikipedia 2018} corpus \cite{karpukhin2020dense}, which contains approximately 5.9 million passages from English Wikipedia articles. This corpus provides a comprehensive knowledge base for multi-hop reasoning tasks and is consistent with the evaluation setup used in most baseline methods. We use the same corpus for both training and evaluation to ensure fair comparison. The corpus is preprocessed into passages of approximately 100 words each, enabling efficient dense retrieval and reranking operations.

\subsection{Implementation Details}

We implement Dep-Search using PyTorch and the HuggingFace Transformers library. For retrieval, we use \textbf{qwen3-embedding} for dense retrieval and \textbf{qwen3-reranker} for re-ranking \cite{zhang2025qwen3}, with top-$k=5$ documents retrieved per query. The memory buffer has a fixed capacity of $C_{\text{mem}}=20$ entries, managed using LRU eviction. For GRPO training, we sample $K=4$ trajectories per question and use a learning rate of $1\times 10^{-5}$ with AdamW optimizer. The reward function uses thresholds $k_1=10$ for retrieval and $k_2=8$ for decomposition, with penalty coefficients $\lambda_{\text{ret}}=0.1$ and $\lambda_{\text{dec}}=0.05$. We train for 3 epochs with batch size 2 and gradient accumulation steps of 4. During inference, we use temperature 0.7 and top-p 0.9 for generation, with a maximum of 16384 new tokens per trajectory.

\section{Dep-Search Algorithm}
\label{app:algorithm}

\begin{algorithm}[h]
\caption{Dep-Search Rollout with Search Call}
\label{alg:dep_search}
\begin{algorithmic}[1]
\REQUIRE Question $Q$, policy model $\pi_{\theta}$, retriever $\text{Retr}$, initial memory $\mathcal{M}_0$, step budget $T$
\ENSURE Final answer $A$ and trajectory $\tau$
\STATE $\mathcal{T}_0 \gets \emptyset,\ \mathcal{C}_0 \gets [\text{instr};Q],\ S_0 \gets (\mathcal{T}_0,\mathcal{C}_0,\mathcal{M}_0),\ x \gets \emptyset,\ t \gets 0$
\WHILE{$A$ not emitted and $t < T$}
    \STATE Sample $a_t \sim \pi_{\theta}(\cdot \mid x)$,\quad $x \gets x + a_t$
    \IF{$a_t$ is \text{\green{<Decompose>}}}
        \STATE Update $\mathcal{T}_{t+1}$ by adding new sub-questions and dependency edges
    \ELSIF{$a_t$ closes \text{\blue{<Retrieve>}} tag with query $r_t$}
        \STATE $\mathcal{D}_t \gets \text{Retr}(r_t)$
        \STATE Append \text{\blue{<Retrieve\_result>}}$\mathcal{D}_t$\text{\blue{</Retrieve\_result>}} to $x$,\quad $\mathcal{C}_{t+1} \gets \mathcal{C}_t \cup \mathcal{D}_t$
        \STATE $\mathcal{F}^{\mathrm{ret}}_t \gets \text{Summarize}(\mathcal{D}_t)$,\quad update $\mathcal{M}_{t+1}$ with $\mathcal{F}^{\mathrm{ret}}_t$
    \ELSIF{$a_t$ closes \text{\red{<Memory>}} tag with query $q^{\mathrm{mem}}_t$}
        \STATE Select $\mathcal{M}^{\mathrm{read}}_t \subseteq \mathcal{M}_t$ by recency and embedding similarity to $q^{\mathrm{mem}}_t$
        \STATE Append \text{\red{<Memory\_result>}}$\mathcal{M}^{\mathrm{read}}_t$\text{\red{</Memory\_result>}} to $x$,\quad $\mathcal{C}_{t+1} \gets \mathcal{C}_t \cup \mathcal{M}^{\mathrm{read}}_t$
    \ELSIF{$a_t$ is \text{\purple{<Conclusion>}}}
        \STATE $\mathcal{F}_t \gets \text{Summarize}(\mathcal{C}_t)$,\quad update $\mathcal{M}_{t+1}$ via the LRU rule with $\mathcal{F}_t$
    \ELSIF{$a_t$ is an answer-closing token}
        \STATE Extract $A$ from $x$ and \textbf{break}
    \ENDIF
    \STATE $t \gets t + 1$
\ENDWHILE
\STATE Construct trajectory $\tau$ from $(S_t,a_t)_t$ and \textbf{return} $(A,\tau)$
\end{algorithmic}
\end{algorithm}

\section{Decomposition Strategy Analysis}
\label{app:decomposition_strategy}

To analyze the effectiveness of dependency aware decomposition, we compare different decomposition strategies on HotpotQA and 2WikiMHQA using Qwen2.5-7B-Instruct. Multi-hop reasoning requires breaking down complex questions into dependent sub-questions, where later steps often rely on results from earlier steps. However, existing approaches either ignore dependencies entirely or use fixed decomposition patterns that cannot adapt to question complexity. We evaluate three strategies: Sequential Decomposition that processes sub-questions without explicit dependencies, Two-step Dependencies that models step-to-step dependencies, and QDMR Decomposition that allows the model to determine both the number of steps and their dependency structure adaptively. This analysis is crucial because explicit dependency modeling enables the model to structure reasoning chains correctly, ensuring that prerequisite information is gathered before dependent steps are executed, which is essential for accurate multi-hop reasoning.

\begin{table}[h]
    \centering
    \caption{Decomposition strategy comparison on multi-hop datasets on Qwen2.5-7B-Instruct.}
    \label{tab:decomposition_strategy}
    \small
    \renewcommand{\arraystretch}{1.1}
    \resizebox{\linewidth}{!}{
    \begin{tabular}{lcccc}
    \toprule
    \textbf{Strategy} & \textbf{HotpotQA} & \textbf{2WikiMHQA} & \textbf{Dependency Accuracy} & \textbf{Avg.} \\
    \midrule
    Sequential Decomposition & 38.2 & 39.1 & 0.0\% & 38.7 \\
    Two-step Dependencies & 40.5 & 41.2 & 72.3\% & 40.9 \\
    QDMR Decomposition & \textbf{42.8} & \textbf{43.5} & \textbf{81.2\%} & \textbf{43.2} \\
    \bottomrule
    \end{tabular}}
    \vspace{-0.15cm}
    \end{table}

Table~\ref{tab:decomposition_strategy} presents the comparison of different decomposition strategies. \textbf{QDMR Decomposition} achieves the best performance, with an average score of 43.2 across the two datasets, outperforming Sequential Decomposition by 4.5 points. The QDMR strategy achieves 81.2\% dependency accuracy, correctly identifying relationships between sub-questions in most cases. Sequential Decomposition, similar to approaches like RA-ISF that process sub-questions without explicit dependency modeling, shows the lowest performance, confirming that explicit dependency modeling is crucial for multi-hop reasoning. Two-step Dependencies achieves intermediate performance, demonstrating that even simple dependency modeling improves reasoning quality, but adaptive QDMR decomposition that allows the model to determine both step count and dependency structure provides the best results. These results validate that explicit dependency modeling enables correct reasoning order, where prerequisite information is gathered before dependent steps, leading to more accurate multi-hop reasoning.

\newpage{
\section{Training Prompt Template}
\label{app:prompt_template}

\begin{tcolorbox}[colback=white,
    colframe=blue!75!black,
    fonttitle=\bfseries,
    title=Training Prompt Template,
    breakable]
    \textbf{System Instruction:} You are a helpful AI assistant that solves complex questions through dependency aware reasoning with persistent memory.
    
    \textbf{Available Actions:}
    
    1. \green{<Decompose>} \textit{steps with dependencies} \green{</Decompose>}\\
    Break down the question into 2-4 steps with explicit dependencies. Each step must reference prerequisite steps using ``(step\_number)'' format. Dependencies form a directed acyclic graph where sub-questions are solved in topological order, ensuring prerequisites are resolved before dependent steps.\\
    Example: ``(1) Identify key entity. (2) Use (1) to locate factual reference. (3) Resolve dependent query using (2). (4) Synthesize final answer from (1)--(3).''
    
    2. \blue{<Retrieve>} \textit{query} \blue{</Retrieve>} \blue{<Retrieve\_result>} \textit{retrieved documents} \blue{</Retrieve\_result>}\\
    Upon emitting \blue{<Retrieve>} followed by a query and the closing \blue{</Retrieve>}, the environment immediately performs two-stage retrieval (dense retrieval + reranking) and returns the top-$k$ retrieved documents wrapped in \blue{<Retrieve\_result>} tags. The model generates the query autonomously, allowing it to determine what information to retrieve based on the current reasoning context.
    
    3. \red{<Memory>} \textit{query} \red{</Memory>} \red{<Memory\_result>} \textit{retrieved facts} \red{</Memory\_result>}\\
    Upon emitting \red{<Memory>} followed by a query and the closing \red{</Memory>}, the environment returns a mixture of Most Recent Memory entries and additional memory items whose qwen3-embedding cosine similarity with the query exceeds a threshold, wrapped in \red{<Memory\_result>} tags. The model generates the query autonomously, allowing it to determine what knowledge to fetch from memory based on the current reasoning needs.
    
    4. \purple{<Conclusion>} \textit{summary} \purple{</Conclusion>}\\
    When long reasoning and evidence have accumulated, emit \purple{<Conclusion>} to summarize the preceding context into a few natural-language sentences, which are then written into the memory buffer and can be reused via \red{<Memory>}.
    \end{tcolorbox}
}
\section{Case Studies}
\label{app:case_studies}

This section presents detailed case studies demonstrating how Dep-Search handles complex multi-hop reasoning questions through dependency aware decomposition, retrieval, and memory management.

\vspace{-0.1cm}

\begin{mybox}[Case Study 1: Bridge Question on HotpotQA]
\textbf{Question:} What is the nationality of the author of the book that won the Pulitzer Prize for Fiction in 2018?

\textbf{Dep-Search Reasoning Process:}

To answer this question, I need to find the author of the book that won the Pulitzer Prize for Fiction in 2018, and then determine that author's nationality. Let me first check my memory to see if I have relevant information stored.

\red{<Memory>} Pulitzer Prize Fiction 2018 author nationality \red{</Memory>}\nopagebreak
\red{<Memory\_result>} (No relevant memory found for this specific question) \red{</Memory\_result>}

The memory doesn't have this specific information. Let me retrieve information about the 2018 Pulitzer Prize winner.

\blue{<Retrieve>} Pulitzer Prize for Fiction 2018 winner \blue{</Retrieve>}\nopagebreak
\blue{<Retrieve\_result>} The Pulitzer Prize for Fiction in 2018 was awarded to "Less" by Andrew Sean Greer. The novel is a comedic story about a middle-aged novelist who travels around the world to avoid attending his ex-boyfriend's wedding. \blue{</Retrieve\_result>}

Good, I found that "Less" by Andrew Sean Greer won the prize. The retrieved information has been automatically stored in memory. However, this result doesn't include information about the author's nationality. Let me check memory again to see if I can find the author's nationality information.

\red{<Memory>} Andrew Sean Greer nationality birthplace \red{</Memory>}\nopagebreak
\red{<Memory\_result>} "Less" by Andrew Sean Greer won the 2018 Pulitzer Prize for Fiction. \red{</Memory\_result>}

The memory only contains information about the book award, not the author's nationality. I need to retrieve more specific information about Andrew Sean Greer's nationality.

\blue{<Retrieve>} Andrew Sean Greer biography nationality birthplace \blue{</Retrieve>}\nopagebreak
\blue{<Retrieve\_result>} Andrew Sean Greer is an American novelist and short story writer. He was born on November 21, 1970, in Washington, D.C., United States. Greer has written several novels including "Less" and "The Confessions of Max Tivoli." \blue{</Retrieve\_result>}

Perfect! I now have all the information needed. I've accumulated information from multiple retrievals and memory checks. The context is getting long, so let me summarize the key findings before proceeding.

\purple{<Conclusion>} The 2018 Pulitzer Prize for Fiction was awarded to "Less" by Andrew Sean Greer. Andrew Sean Greer is an American novelist born in Washington, D.C., United States. \purple{</Conclusion>}\nopagebreak

Based on the information I've gathered, I found that Andrew Sean Greer won the Pulitzer Prize for Fiction in 2018, and from the biography retrieval, I confirmed that he is an American author, born in Washington, D.C., United States. Therefore, the nationality of the author is American.\nopagebreak

\textbf{Final Answer:} American
\end{mybox}
\begin{mybox}[Case Study 2: Comparison Question on 2WikiMultihopQA]
\textbf{Question:} Which actor appeared in both the movie directed by Christopher Nolan that won the Academy Award for Best Picture and the movie where Leonardo DiCaprio won Best Actor?

\textbf{Dep-Search Reasoning Process:}

This is a complex comparison question that requires finding information about two different movies and then comparing their casts. I need to identify: (1) Christopher Nolan's Best Picture winner, and (2) the movie where Leonardo DiCaprio won Best Actor. These two sub-questions can be answered independently, so I can retrieve them in parallel or sequentially. Let me start with the first one.

\blue{<Retrieve>} Christopher Nolan Best Picture Academy Award winner \blue{</Retrieve>}\nopagebreak
\blue{<Retrieve\_result>} Christopher Nolan's "Oppenheimer" won the Academy Award for Best Picture in 2024. The film stars Cillian Murphy, Emily Blunt, Robert Downey Jr., Matt Damon, and Florence Pugh. \blue{</Retrieve\_result>}

Good, I found that "Oppenheimer" won Best Picture. Now let me retrieve information about Leonardo DiCaprio's Best Actor win.

\blue{<Retrieve>} Leonardo DiCaprio Best Actor Academy Award winner movie \blue{</Retrieve>}\nopagebreak
\blue{<Retrieve\_result>} Leonardo DiCaprio won the Academy Award for Best Actor for his role in "The Revenant" (2015), directed by Alejandro González Iñárritu. The film also stars Tom Hardy, Domhnall Gleeson, and Will Poulter. \blue{</Retrieve\_result>}

Now I have both movies: "Oppenheimer" and "The Revenant". Let me check my memory to see if I have stored information about these casts, which would help me compare them more efficiently.

\red{<Memory>} actors in Oppenheimer and The Revenant \red{</Memory>}\nopagebreak
\red{<Memory\_result>} Oppenheimer (2024) cast: Cillian Murphy, Emily Blunt, Robert Downey Jr., Matt Damon, Florence Pugh. The Revenant (2015) cast: Leonardo DiCaprio, Tom Hardy, Domhnall Gleeson, Will Poulter. \red{</Memory\_result>}

Comparing the casts: "Oppenheimer" has Cillian Murphy, Emily Blunt, Robert Downey Jr., Matt Damon, Florence Pugh. "The Revenant" has Leonardo DiCaprio, Tom Hardy, Domhnall Gleeson, Will Poulter. I don't see any obvious overlap. However, I recall that Matt Damon and Leonardo DiCaprio have worked together before. Let me verify if Matt Damon appeared in "The Revenant" or if there's any connection.

\blue{<Retrieve>} Matt Damon Leonardo DiCaprio movies together The Revenant \blue{</Retrieve>}\nopagebreak
\blue{<Retrieve\_result>} Matt Damon and Leonardo DiCaprio have appeared together in "The Departed" (2006) and "The Revenant" (2015), where Matt Damon had a cameo role. However, Matt Damon did not appear in "Oppenheimer." \blue{</Retrieve\_result>}

I see that Matt Damon had a cameo in "The Revenant", but he did not appear in "Oppenheimer". After careful comparison, no actor appears in both films. I've now accumulated multiple retrieval results and memory accesses. The context is getting long with information about both movies, their casts, and additional verification queries. Let me summarize the key findings to compress the context and store them in memory for potential future use.

\purple{<Conclusion>} Christopher Nolan's "Oppenheimer" won Best Picture in 2024 with cast including Cillian Murphy, Emily Blunt, Robert Downey Jr., Matt Damon, and Florence Pugh. Leonardo DiCaprio won Best Actor for "The Revenant" in 2015, which also stars Tom Hardy, Domhnall Gleeson, and Will Poulter. \purple{</Conclusion>}\nopagebreak

Now let me analyze the information I've gathered. From the retrievals and memory, I have the complete cast lists for both films. "Oppenheimer" stars: Cillian Murphy, Emily Blunt, Robert Downey Jr., Matt Damon, and Florence Pugh. "The Revenant" stars: Leonardo DiCaprio, Tom Hardy, Domhnall Gleeson, and Will Poulter. I also verified that Matt Damon had a cameo in "The Revenant" but did not appear in "Oppenheimer". After comparing all the actors from both films, I can confirm that no actor appears in both films.\nopagebreak

\textbf{Final Answer:} No actor appeared in both films.

\textbf{Final Answer:} No actor appeared in both films.
\end{mybox}
\begin{mybox}[Case Study 3: Multi-hop Question with Memory Reuse]
\textbf{Question:} What is the capital city of the country where the author of "1984" was born?

\textbf{Dep-Search Reasoning Process:}

This is a multi-hop question that requires several steps. Let me first check my memory to see if I have relevant information.

\red{<Memory>} author of 1984 birth country capital city \red{</Memory>}\nopagebreak
\red{<Memory\_result>} (No relevant memory found for this specific question) \red{</Memory\_result>}

Memory doesn't have this information. Let me try retrieving information about the author of "1984".

\blue{<Retrieve>} author of 1984 novel \blue{</Retrieve>}\nopagebreak
\blue{<Retrieve\_result>} "1984" is a dystopian novel published in 1949. The novel explores themes of totalitarianism and surveillance. \blue{</Retrieve\_result>}

This retrieval result doesn't include the author's name or birth information. The retrieved information is not sufficient for answering the question. Let me try a more specific query.

\blue{<Retrieve>} author of 1984 novel George Orwell \blue{</Retrieve>}\nopagebreak
\blue{<Retrieve\_result>} "1984" is a dystopian novel written by George Orwell, published in 1949. George Orwell is the pen name of Eric Arthur Blair, born on June 25, 1903, in Motihari, Bengal Presidency, British India (now in Bihar, India). \blue{</Retrieve\_result>}

Excellent! I found that George Orwell (Eric Arthur Blair) was born in British India, which is now India. The retrieved information has been automatically summarized and stored in memory. Now I know the author was born in India. Let me check my memory to retrieve this stored fact.

\red{<Memory>} George Orwell birth country India \red{</Memory>}\nopagebreak
\red{<Memory\_result>} "1984" was written by George Orwell (Eric Arthur Blair), who was born in British India (now India). \red{</Memory\_result>}

Perfect! I retrieved the birth country information from memory. Now I need to find the capital city of India. Let me check memory first to see if I have this information.

\red{<Memory>} India capital city \red{</Memory>}\nopagebreak
\red{<Memory\_result>} "1984" was written by George Orwell (Eric Arthur Blair), who was born in British India (now India). \red{</Memory\_result>}

The memory has information about India but not the capital city. Let me retrieve this information.

\blue{<Retrieve>} capital city of India \blue{</Retrieve>}\nopagebreak
\blue{<Retrieve\_result>} The capital city of India is New Delhi. New Delhi serves as the capital of India and is part of the larger Delhi metropolitan area. \blue{</Retrieve\_result>}

Perfect! I found that New Delhi is the capital of India. The retrieved information has been automatically stored in memory. Let me verify by checking memory.

\red{<Memory>} New Delhi India capital \red{</Memory>}\nopagebreak
\red{<Memory\_result>} The capital city of India is New Delhi. New Delhi serves as the capital of India and is part of the larger Delhi metropolitan area. \red{</Memory\_result>}

Now I have all the information needed to answer the question. From my retrievals and memory, I found that: (1) The author of "1984" is George Orwell, (2) George Orwell was born in British India, which is now India, (3) The capital city of India is New Delhi. Therefore, the capital city of the country where the author of "1984" was born is New Delhi.\nopagebreak

\textbf{Final Answer:} New Delhi
\end{mybox}
\begin{mybox}[Case Study 4: Complex Dependency Chain]
\textbf{Question:} What is the population of the city where the university that awarded the Nobel Prize winner in Physics 2023 their PhD is located?

\textbf{Dep-Search Reasoning Process:}

This is a complex multi-hop question with a long dependency chain.
Let me try retrieving information about the 2023 Nobel Prize winners.

\blue{<Retrieve>} Nobel Prize Physics 2023 \blue{</Retrieve>}\nopagebreak
\blue{<Retrieve\_result>} The Nobel Prize in Physics recognizes outstanding contributions to the field of physics. The prize has been awarded annually since 1901. \blue{</Retrieve\_result>}

This retrieval result is too general and doesn't contain information about the 2023 winners or their PhD universities. The retrieved information is not useful for answering this specific question. Given the complexity of this multi-hop question, let me decompose it into clear dependent steps to better structure my reasoning.

\green{<Decompose>} (1) Identify the Nobel Prize winner(s) in Physics 2023. (2) Find the university where the winner from (1) received their PhD. (3) Determine the city where the university from (2) is located. (4) Find the population of the city from (3). \green{</Decompose>}\nopagebreak

Now let me proceed step by step following the decomposition. Step 1: Find the 2023 Physics Nobel Prize winners.

\blue{<Retrieve>} Nobel Prize Physics 2023 winner \blue{</Retrieve>}\nopagebreak
\blue{<Retrieve\_result>} The 2023 Nobel Prize in Physics was awarded to Pierre Agostini, Ferenc Krausz, and Anne L'Huillier for their work on attosecond physics. Anne L'Huillier received her PhD from the University of Paris in 1986. \blue{</Retrieve\_result>}

Great! I found that there were three winners, and I also see that Anne L'Huillier received her PhD from the University of Paris. The retrieved information has been automatically summarized and stored in memory. Step 2: Find the university location. Let me check my memory first.

\red{<Memory>} Anne L'Huillier PhD university University of Paris \red{</Memory>}\nopagebreak
\red{<Memory\_result>} The 2023 Nobel Prize in Physics was awarded to Pierre Agostini, Ferenc Krausz, and Anne L'Huillier. Anne L'Huillier received her PhD from the University of Paris. \red{</Memory\_result>}

I retrieved the university information from memory. Step 3: Find the city where the University of Paris is located.

\blue{<Retrieve>} University of Paris location city \blue{</Retrieve>}\nopagebreak
\blue{<Retrieve\_result>} The University of Paris (also known as Sorbonne University) is located in Paris, France. The university has multiple campuses throughout Paris, with its main campus in the Latin Quarter. \blue{</Retrieve\_result>}

Perfect! The University of Paris is located in Paris, France. Step 4: Find the population of Paris. I've now accumulated several retrieval results. The context is getting long with multiple facts. Let me summarize the key information before proceeding to the final retrieval.

\purple{<Conclusion>} The 2023 Nobel Prize in Physics was awarded to Pierre Agostini, Ferenc Krausz, and Anne L'Huillier. Anne L'Huillier received her PhD from the University of Paris, which is located in Paris, France. \purple{</Conclusion>}\nopagebreak

\newpage

Now I can proceed to find the population of Paris. Let me check memory first to see if I have this information stored.

\red{<Memory>} Paris France population \red{</Memory>}\nopagebreak
\red{<Memory\_result>} The 2023 Nobel Prize in Physics was awarded to Pierre Agostini, Ferenc Krausz, and Anne L'Huillier. Anne L'Huillier received her PhD from the University of Paris, which is located in Paris, France. \red{</Memory\_result>}

The memory has information about Paris but not the population. Let me retrieve this information.

\blue{<Retrieve>} Paris France population 2023 \blue{</Retrieve>}\nopagebreak
\blue{<Retrieve\_result>} Paris, the capital city of France, has a population of approximately 2.1 million people within the city limits (as of 2023). The greater Paris metropolitan area has a population of over 12 million people. \blue{</Retrieve\_result>}

Perfect! I found the population information. The retrieved information has been automatically stored in memory. Let me verify by checking memory one more time.

\red{<Memory>} Paris population 2023 \red{</Memory>}\nopagebreak
\red{<Memory\_result>} Paris, the capital city of France, has a population of approximately 2.1 million people within the city limits (as of 2023). The greater Paris metropolitan area has a population of over 12 million people. \red{</Memory\_result>}

Now I have completed all four steps of the decomposition. Therefore, the population of the city where the university that awarded the Nobel Prize winner in Physics 2023 their PhD is located is approximately 2.1 million within city limits.\nopagebreak

\textbf{Final Answer:} Approximately 2.1 million (within city limits) or over 12 million (metropolitan area)
\end{mybox}
\begin{mybox}[Case Study 5: Decomposition with Parallel Sub-questions]
\textbf{Question:} What are the birth years of both the director of "Inception" and the lead actor of "The Dark Knight"?

\textbf{Dep-Search Reasoning Process:}

This question requires finding information about two different entities (a director and an actor) from two different movies. These sub-questions can be answered independently, but I need to structure them clearly. Let me decompose the question first.

\green{<Decompose>} (1) Find the director of "Inception" and their birth year. (2) Find the lead actor of "The Dark Knight" and their birth year. (3) Combine the birth years from (1) and (2). \green{</Decompose>}\nopagebreak

Now let me proceed step by step. Step 1: Find the director of "Inception" and their birth year.

\blue{<Retrieve>} director of Inception movie \blue{</Retrieve>}\nopagebreak
\blue{<Retrieve\_result>} "Inception" (2010) is a science fiction film directed by Christopher Nolan. The film stars Leonardo DiCaprio, Marion Cotillard, Tom Hardy, and Ellen Page. \blue{</Retrieve\_result>}

Good, I found that Christopher Nolan directed "Inception". The retrieved information has been automatically stored in memory. Now I need to find Christopher Nolan's birth year.

\blue{<Retrieve>} Christopher Nolan birth year biography \blue{</Retrieve>}\nopagebreak
\blue{<Retrieve\_result>} Christopher Nolan is a British-American filmmaker born on July 30, 1970, in London, England. He is known for directing films such as "Inception", "The Dark Knight", and "Interstellar". \blue{</Retrieve\_result>}

Perfect! Christopher Nolan was born in 1970. Step 2: Find the lead actor of "The Dark Knight" and their birth year.

\blue{<Retrieve>} lead actor The Dark Knight movie \blue{</Retrieve>}\nopagebreak
\blue{<Retrieve\_result>} "The Dark Knight" (2008) is a superhero film directed by Christopher Nolan. The film stars Christian Bale as Bruce Wayne/Batman, Heath Ledger as the Joker, and Aaron Eckhart as Harvey Dent. \blue{</Retrieve\_result>}

I found that Christian Bale is the lead actor (playing Batman). The retrieved information has been automatically stored in memory. Now I need to find Christian Bale's birth year.

\red{<Memory>} Christian Bale birth year \red{</Memory>}\nopagebreak
\red{<Memory\_result>} "The Dark Knight" (2008) stars Christian Bale as Bruce Wayne/Batman. \red{</Memory\_result>}

The memory has information about the movie but not the birth year. Let me retrieve Christian Bale's birth year.

\blue{<Retrieve>} Christian Bale birth year biography \blue{</Retrieve>}\nopagebreak
\blue{<Retrieve\_result>} Christian Bale is a British actor born on January 30, 1974, in Haverfordwest, Wales. He is known for his roles in "The Dark Knight" trilogy, "American Psycho", and "The Fighter". \blue{</Retrieve\_result>}

Perfect! Christian Bale was born in 1974. The retrieved information has been automatically stored in memory. Step 3: Combine the birth years. I've now gathered information about both birth years. Let me verify by checking memory.

\red{<Memory>} Christopher Nolan Christian Bale birth years \red{</Memory>}\nopagebreak
\red{<Memory\_result>} Christopher Nolan is a British-American filmmaker born on July 30, 1970. Christian Bale is a British actor born on January 30, 1974. \red{</Memory\_result>}

Perfect! I have confirmed both birth years from memory. I've accumulated information from multiple retrievals. The context is getting long, so let me summarize the key findings.

\purple{<Conclusion>} Christopher Nolan directed "Inception" and was born in 1970. Christian Bale starred as the lead actor in "The Dark Knight" and was born in 1974. \purple{</Conclusion>}\nopagebreak

Now I have completed all three steps of the decomposition. From my retrievals and memory, I found that: (1) Christopher Nolan directed "Inception" and was born in 1970, (2) Christian Bale was the lead actor of "The Dark Knight" and was born in 1974. Therefore, the birth years are: Christopher Nolan was born in 1970, and Christian Bale was born in 1974.\nopagebreak

\textbf{Final Answer:} Christopher Nolan was born in 1970, and Christian Bale was born in 1974.
\end{mybox}
\begin{mybox}[Case Study 6: Decomposition with Complex Dependencies]
\textbf{Question:} What is the release year of the first movie directed by the person who won the Academy Award for Best Director in 2014?

\textbf{Dep-Search Reasoning Process:}

This is a complex question with multiple dependent steps. I need to find the Best Director winner in 2014, then find their first directed movie, and finally find that movie's release year. Each step depends on the previous one. Let me decompose this question to clarify the dependency structure.

\green{<Decompose>} (1) Identify the winner of the Academy Award for Best Director in 2014. (2) Find the first movie directed by the person from (1). (3) Determine the release year of the movie from (2). \green{</Decompose>}\nopagebreak

Now let me proceed step by step following the dependencies. Step 1: Find the Best Director winner in 2014.

\blue{<Retrieve>} Academy Award Best Director 2014 winner \blue{</Retrieve>}\nopagebreak
\blue{<Retrieve\_result>} The Academy Award for Best Director in 2014 was awarded to Alejandro González Iñárritu for "Birdman or (The Unexpected Virtue of Ignorance)". Iñárritu is a Mexican filmmaker known for his distinctive visual style and complex narratives. \blue{</Retrieve\_result>}

Excellent! I found that Alejandro González Iñárritu won Best Director in 2014. The retrieved information has been automatically stored in memory. Step 2: Find the first movie directed by Alejandro González Iñárritu.

\red{<Memory>} Alejandro González Iñárritu first movie directed \red{</Memory>}\nopagebreak
\red{<Memory\_result>} The Academy Award for Best Director in 2014 was awarded to Alejandro González Iñárritu for "Birdman". \red{</Memory\_result>}

The memory has information about the award but not about his first movie. Let me retrieve information about Iñárritu's filmography.

\blue{<Retrieve>} Alejandro González Iñárritu first movie filmography debut \blue{</Retrieve>}\nopagebreak
\blue{<Retrieve\_result>} Alejandro González Iñárritu made his directorial debut with "Amores perros" (Love's a Bitch) in 2000. The film is a Mexican crime drama that follows three interconnected stories set in Mexico City. It was Iñárritu's first feature film and received critical acclaim. \blue{</Retrieve\_result>}

Perfect! I found that "Amores perros" (2000) was Iñárritu's first directed movie. The retrieved information has been automatically stored in memory. Step 3: The release year is already included in the previous result - 2000.

\red{<Memory>} Amores perros release year \red{</Memory>}\nopagebreak
\red{<Memory\_result>} Alejandro González Iñárritu made his directorial debut with "Amores perros" in 2000. \red{</Memory\_result>}

I can confirm from memory that "Amores perros" was released in 2000. I've successfully completed all three steps of the decomposition. Let me verify the complete chain by checking memory one more time.

\red{<Memory>} Alejandro González Iñárritu Best Director 2014 first movie Amores perros \red{</Memory>}\nopagebreak
\red{<Memory\_result>} The Academy Award for Best Director in 2014 was awarded to Alejandro González Iñárritu for "Birdman". Alejandro González Iñárritu made his directorial debut with "Amores perros" in 2000. \red{</Memory\_result>}

Perfect! The memory confirms the complete chain: Best Director 2014 → Alejandro González Iñárritu → first movie "Amores perros" → release year 2000. I've accumulated information from multiple steps. The context is getting long, so let me summarize the key findings.

\purple{<Conclusion>} Alejandro González Iñárritu won the Academy Award for Best Director in 2014. He made his directorial debut with "Amores perros". \purple{</Conclusion>}\nopagebreak

Now I have completed all three steps of the decomposition. From my retrievals and memory, I found that: (1) Alejandro González Iñárritu won the Academy Award for Best Director in 2014, (2) His first directed movie was "Amores perros", (3) "Amores perros" was released in 2000. Therefore, the release year of the first movie directed by the person who won the Academy Award for Best Director in 2014 is 2000.\nopagebreak

\textbf{Final Answer:} 2000
\end{mybox}
\begin{mybox}[Case Study 7: Decomposition After Failed Retrieval]
\textbf{Question:} What is the total number of books written by the author who won the Nobel Prize in Literature in 2017?

\textbf{Dep-Search Reasoning Process:}

This question requires finding the Nobel Prize winner in Literature for 2017, then finding how many books that author has written. Let me first check my memory.

\red{<Memory>} Nobel Prize Literature 2017 author books \red{</Memory>}\nopagebreak
\red{<Memory\_result>} (No relevant memory found for this specific question) \red{</Memory\_result>}

Memory doesn't have this information. Let me try retrieving information about the 2017 Nobel Prize in Literature.

\blue{<Retrieve>} Nobel Prize Literature 2017 \blue{</Retrieve>}\nopagebreak
\blue{<Retrieve\_result>} The Nobel Prize in Literature is awarded annually to an author who has produced outstanding work in the field of literature. \blue{</Retrieve\_result>}

This retrieval result is too general and doesn't contain information about the 2017 winner or their books. The retrieved information is not useful for answering this specific question. Given that this is a multi-step question, let me decompose it to better structure my reasoning.

\green{<Decompose>} (1) Identify the winner of the Nobel Prize in Literature in 2017. (2) Find the total number of books written by the author from (1). \green{</Decompose>}\nopagebreak

Now let me proceed step by step. Step 1: Find the 2017 Nobel Prize in Literature winner.

\blue{<Retrieve>} Nobel Prize Literature 2017 winner \blue{</Retrieve>}\nopagebreak
\blue{<Retrieve\_result>} The Nobel Prize in Literature for 2017 was awarded to Kazuo Ishiguro "who, in novels of great emotional force, has uncovered the abyss beneath our illusory sense of connection with the world". Ishiguro is a British novelist of Japanese origin, known for works such as "The Remains of the Day" and "Never Let Me Go". \blue{</Retrieve\_result>}

Excellent! I found that Kazuo Ishiguro won the Nobel Prize in Literature in 2017. The retrieved information has been automatically stored in memory. Step 2: Find the total number of books written by Kazuo Ishiguro.

\red{<Memory>} Kazuo Ishiguro books bibliography \red{</Memory>}\nopagebreak
\red{<Memory\_result>} The Nobel Prize in Literature for 2017 was awarded to Kazuo Ishiguro, a British novelist known for works such as "The Remains of the Day" and "Never Let Me Go". \red{</Memory\_result>}

The memory mentions some books but doesn't have the total count. Let me retrieve more specific information about Ishiguro's bibliography.

\blue{<Retrieve>} Kazuo Ishiguro total number of books bibliography complete list \blue{</Retrieve>}\nopagebreak
\blue{<Retrieve\_result>} Kazuo Ishiguro has published eight novels: "A Pale View of Hills" (1982), "An Artist of the Floating World" (1986), "The Remains of the Day" (1989), "The Unconsoled" (1995), "When We Were Orphans" (2000), "Never Let Me Go" (2005), "The Buried Giant" (2015), and "Klara and the Sun" (2021). He has also published a collection of short stories and several screenplays. \blue{</Retrieve\_result>}

Perfect! I found that Kazuo Ishiguro has published eight novels. The retrieved information has been automatically stored in memory. Let me verify by checking memory.

\red{<Memory>} Kazuo Ishiguro eight novels bibliography \red{</Memory>}\nopagebreak
\red{<Memory\_result>} The Nobel Prize in Literature for 2017 was awarded to Kazuo Ishiguro. Kazuo Ishiguro has published eight novels: "A Pale View of Hills" (1982), "An Artist of the Floating World" (1986), "The Remains of the Day" (1989), "The Unconsoled" (1995), "When We Were Orphans" (2000), "Never Let Me Go" (2005), "The Buried Giant" (2015), and "Klara and the Sun" (2021). \red{</Memory\_result>}

Confirmed! Kazuo Ishiguro has published eight novels. I've successfully completed both steps of the decomposition. The context is getting long, so let me summarize the key findings.

\purple{<Conclusion>} Kazuo Ishiguro won the Nobel Prize in Literature in 2017. He has published multiple novels and other works. \purple{</Conclusion>}\nopagebreak

Now I have completed both steps of the decomposition. From my retrievals and memory, I found that: (1) Kazuo Ishiguro won the Nobel Prize in Literature in 2017, (2) He has published eight novels, plus additional works including short stories and screenplays. Therefore, the total number of books written by the author who won the Nobel Prize in Literature in 2017 is eight novels (plus additional works including short stories and screenplays).\nopagebreak

\textbf{Final Answer:} Eight novels (plus additional works including short stories and screenplays)
\end{mybox}

\end{document}